\documentclass{article}

\PassOptionsToPackage{numbers, compress}{natbib}

\usepackage[preprint]{neurips_2026}

\usepackage[utf8]{inputenc} 
\usepackage[T1]{fontenc}    
\usepackage{hyperref}       
\usepackage{url}            
\usepackage{booktabs}       
\usepackage{amsmath}        
\usepackage{amssymb}        
\usepackage{amsfonts}       
\usepackage{bm}             
\usepackage{nicefrac}       
\usepackage{microtype}      
\usepackage[table]{xcolor}  
\usepackage{graphicx}       
\usepackage{adjustbox}
\usepackage{makecell}       
\usepackage{multirow}       
\usepackage{wrapfig}        
\usepackage{enumitem}       
\usepackage{titletoc}
\title{FlowWAM: Optical Flow as a Unified Action Representation for World Action Models}

\author{
\textbf{Yixiang~Chen}\textsuperscript{\normalfont 1,2,$\ast$} \quad
\textbf{Peiyan~Li}\textsuperscript{\normalfont 1,2,$\ast$} \quad
\textbf{Yuan~Xu}\textsuperscript{\normalfont 1,2} \quad
\textbf{Qisen~Ma}\textsuperscript{\normalfont 1,2} \\
\textbf{Jiabing~Yang}\textsuperscript{\normalfont 1,2} \quad
\textbf{Kai~Wang}\textsuperscript{\normalfont 1,2} \quad
\textbf{Jianhua~Yang}\textsuperscript{\normalfont 1,2} \quad
\textbf{Dong~An}\textsuperscript{\normalfont 4} \\
\textbf{He~Guan}\textsuperscript{\normalfont 3} \quad
\textbf{Gaoteng~Liu}\textsuperscript{\normalfont 3} \quad
\textbf{Jianlou~Si}\textsuperscript{\normalfont 5} \quad
\textbf{Jun~Huang}\textsuperscript{\normalfont 5} \\
\textbf{Jing~Liu}\textsuperscript{\normalfont 3} \quad
\textbf{Nianfeng~Liu}\textsuperscript{\normalfont 3} \quad
\textbf{Yan~Huang}\textsuperscript{\normalfont 1,2,3,$\dagger$} \quad
\textbf{Liang~Wang}\textsuperscript{\normalfont 1,2,$\dagger$} \\
\textsuperscript{1}New Laboratory of Pattern Recognition (NLPR), \\ Institute of Automation, Chinese Academy of Sciences \\
\textsuperscript{2}School of Artificial Intelligence, University of Chinese Academy of Sciences \\
\textsuperscript{3}FiveAges \quad
\textsuperscript{4}MBZUAI \quad
\textsuperscript{5}Alibaba Group \\
{\tt\small yixiang.chen@cripac.ia.ac.cn \quad \{yhuang, wangliang\}@nlpr.ia.ac.cn}
}

\begin{document}

\maketitle
\begingroup
\renewcommand{\thefootnote}{\fnsymbol{footnote}}
\footnotetext[1]{Equal contribution.}
\footnotetext[2]{Corresponding author.}
\endgroup

\begin{abstract}
World Action Models (WAMs) are able to leverage pretrained video generators for both world modeling and action prediction.
However, directly leveraging such video generators for control raises a new challenge: how to represent actions in a suitable form that aligns with pretrained video generators while carrying enough motion cues for accurate control.
Existing numerical actions fail to satisfy the former, and prior visual action representations overlook the temporal motion structure across frames.
We address this issue with \textbf{FlowWAM}, a dual-stream diffusion framework that adopts optical flow as a unified, video-native action representation. Flow videos share the same format as RGB videos and encode rich per-pixel displacement. By jointly modeling them within a shared pretrained video generator, FlowWAM can naturally implement two modes of WAMs. In policy mode, FlowWAM generates flow for action prediction, while in world-model mode, it uses target flow sequences to guide future video generation.
Moreover, since flow can be easily extracted from raw videos without action labels, FlowWAM can leverage large-scale action-unlabeled video datasets for pretraining.
We empirically find that our flow-based action representation delivers gains across both modes. On RoboTwin manipulation, FlowWAM raises the success rate to \textbf{92.94\%} on the \emph{Clean} setting and \textbf{92.14\%} on \emph{Random}, outperforming both VLA and WAM baselines. On WorldArena world modeling, it achieves the best overall EWMScore (\textbf{63.71}) with an \textbf{18.4\%} relative improvement in trajectory accuracy. More results can be found on our project
website: \href{https://flow-wam.github.io/}{https://flow-wam.github.io}.
\end{abstract}
\section{Introduction}
\begin{figure*}[t]
  \centering
  \includegraphics[width=0.95\linewidth]{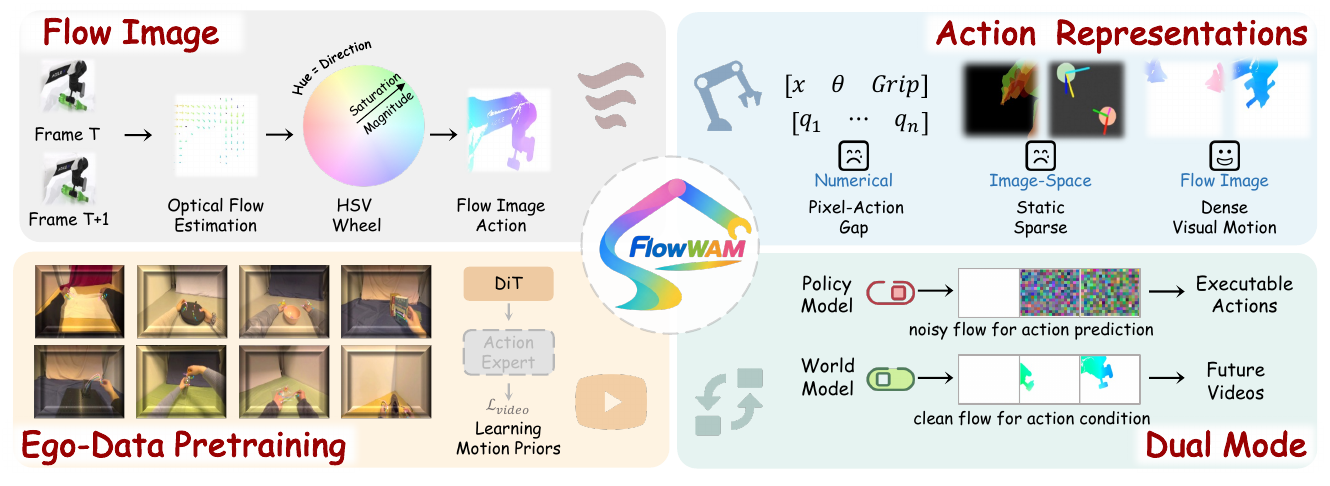}
  \caption{
    \textbf{FlowWAM uses optical flow as a unified action representation for world action models.}
    By representing actions with dense flow videos, FlowWAM bridges executable robot actions and pixel-space video priors, enabling action-unlabeled video pretraining, policy inference, and action-conditioned world modeling within the same framework.
  }
  \label{fig:teaser}
  \vspace{-0.4cm}
\end{figure*}

World Action Models (WAMs)~\citep{dreamzero, uwm, covar, motus, cosmospolicy, mimicvideo, lingbot, fast-wam} have recently emerged as a promising direction alongside Vision-Language-Action (VLA) policies~\citep{pi_0, openvla-oft, bridgevla} for manipulation.
While VLA policies often benefit from semantic priors in image-language pretraining, WAMs leverage pretrained video generators that already encode motion and temporal dynamics.
These motion-centric priors are particularly well-suited for manipulation, where success depends not only on grounding instructions to objects but also on capturing the continuous, frame-by-frame motion of both the robot and the environment.
To use these video priors for control, an ideal action representation should fit the visual input format of pretrained video generators while preserving the cross-frame motion cues needed for control.

Existing WAMs have explored several forms of action representation, yielding substantial advancements in the field.
Numerical action tokens~\citep{dreamzero, lingbot, mimicvideo, fast-wam} are precise, but different robots use different action spaces, making direct transfer across embodiments difficult.
Learned latent actions~\citep{motus, lapa} reduce this dependence by learning from frame transitions, yet they are abstract and may not preserve the dense, spatially grounded motion cues needed for control.
Recent image-space actions, such as action ray maps~\citep{evac}, masks~\citep{bridgev2w}, or multi-view action images~\citep{mv-vdp} reduce this gap by rendering control signals into visual form.
However, these visual conditions are largely static spatial cues that indicate where actions occur, rather than how each visible part of the embodiment moves across frames.
As a result, they remain frame-level conditioning signals injected into the video generator, rather than a temporally dense action representation that evolves jointly with the predicted future videos.
These limitations leave an open question: \textbf{how can actions be represented in a video-native form that preserves dense cross-frame motion cues while remaining decodable into executable control?}

Our key observation is that \textbf{optical flow serves as a naturally suitable action representation for WAMs}.
By recording per-pixel displacement and mapping it to the RGB space via HSV encoding, flow videos become dense, video-like motion signals that align seamlessly with the visual priors of pretrained video generators.
Furthermore, because flow can be extracted directly from raw egocentric videos, it provides a unified representation for both action-unlabeled pretraining and downstream control.
While prior works treat flow as an auxiliary signal \citep{vladm, flowvla} or a modular intermediate~\citep{im2flow2act, fofpred}, we instead re-envision optical flow as a primary, video-native action modality that is deeply integrated into the WAM's generative latent space. Despite its potential, this direction remains largely underexplored.

Building on these insights, we propose \textbf{FlowWAM}, a flow-based dual-stream diffusion framework that jointly integrates action prediction and world modeling within a unified diffusion backbone. By leveraging the fact that flow shares the same image format as RGB frames, FlowWAM enables both streams to share the same VAE encoder and transformer blocks, with only lightweight stream-specific adapters. This architecture effectively casts both action prediction and world modeling as a single, consistent video-to-video generation task, allowing the model to directly inherit the rich motion priors of large-scale video generators. As illustrated in Fig.~\ref{fig:teaser}, FlowWAM establishes flow as a \textbf{unified action representation}. In policy mode, the model generates future flow that is decoded into executable actions, while in world-model mode, it uses target flow sequences to guide video generation. FlowWAM offers three key technical advantages:
(1) Uniformity: flow shares the same image format as RGB, closing the modality gap between control signals and pretrained video priors.
(2) Reliability: dense flow provides a spatially grounded conditioning signal for video prediction, making action-conditioned world modeling more faithful and reliable than conditioning on sparse action tokens.
(3) Scalability: flow can be extracted from egocentric videos without action labels, allowing FlowWAM to acquire motion priors from large-scale unlabeled video sources.

Our contributions are summarized as follows:
\begin{itemize}
  \item To our knowledge, we are the first to introduce optical flow as a unified action representation for WAMs, providing a dense, video-native motion signal that can be predicted, conditioned on, and learned from action-unlabeled egocentric videos.
  \item We propose FlowWAM, a dual-stream diffusion framework that jointly models RGB and flow videos with a shared pretrained video generator, enabling both action prediction for manipulation and video generation for world modeling.
  \item We demonstrate FlowWAM in both policy and world-model settings: on RoboTwin~\citep{robotwin2}, FlowWAM achieves \textbf{92.94\%} success on \emph{Clean} and \textbf{92.14\%} on \emph{Random}, outperforming both VLA and WAM baselines. On WorldArena~\citep{worldarena}, FlowWAM achieves a state-of-the-art EWMScore of \textbf{63.71}, with significant gains in action-conditioned trajectory modeling.
\end{itemize}
\section{Related Work}
\label{sec:related}

\paragraph{World Action Models.}
Recent World Action Models (WAMs) adapt video generators for robot manipulation by coupling action prediction with world modeling. To allow the imagined future to guide policy learning,
one line jointly generates videos and actions within a shared generative process. For example, DreamZero~\citep{dreamzero} and Cosmos Policy~\citep{cosmospolicy} fine-tune video generators with action tokens, while UWM~\citep{uwm} and CoVAR~\citep{covar} couple video and action diffusion heads through shared latents.
Another line follows a generate-then-infer pipeline, first producing future videos and then recovering actions with an inverse dynamics model~\citep{vpp,mimicvideo,vidarc} or causal prediction module~\citep{lingbot}.
These methods demonstrate the value of video priors for control, but actions usually remain a heterogeneous stream that must be connected to video latents through action-specific tokens, heads, or decoders.
FlowWAM differs by representing action-relevant motion itself as a flow video, allowing the action representation to be modeled as a visual stream inside the same pretrained video generator.
This keeps the action representation close to the motion patterns already captured by the video prior.

\paragraph{Action Representations in WAMs.}
Existing unified WAMs can be grouped by how the action stream relates to the video generator.
One line uses numerical action tokens~\citep{dreamzero,cosmospolicy,uwm} that are appended to or co-generated with video latents, leveraging the video generator for prediction while keeping actions in a separate symbolic space.
Another line learns latent action codes from frame transitions~\citep{lapa,motus,lawm}, providing an action representation that is independent of any specific robot.
A third line conditions video generation on image-space action signals such as ray maps~\citep{evac}, embodiment masks~\citep{bridgev2w}, or multi-view action images~\citep{mv-vdp}, placing action information directly in the input domain of the video generator.
FlowWAM falls into this third line, but replaces sparse spatial cues with dense optical-flow videos. This single video-formatted stream acts as both the action prediction target in policy mode and the conditioning signal in world-model mode.

\paragraph{Flow for Manipulation.}
Optical flow and pixel motion have also been used as action representations for manipulation, especially when policies need explicit visual motion cues.
Prior work uses flow as an auxiliary supervision signal or motion target for VLA-style policies~\citep{vladm,flowvla,motus}, or as an intermediate variable in flow-to-action, planning, and future-flow prediction pipelines~\citep{im2flow2act,ecflow,flip,fofpred}.
Recent pixel-motion systems such as LangToMo~\citep{langtomo} and DAWN~\citep{dawn} similarly bridge high-level motion generation and low-level control through structured pixel motion, while trajectory methods such as ATM~\citep{atm} model sparse point tracks.
These methods demonstrate the value of motion cues, but they restrictively keep flow outside the video generator as an auxiliary target, planning variable, or bridge between modules. FlowWAM instead models dense flow as a native stream that can be generated for policy inference and used to condition action-sensitive world modeling.
The same motion representation supports both control and prediction within one video generator.

\section{Method}
\label{sec:method}

\begin{figure*}[t]
  \centering
  \includegraphics[width=\linewidth]{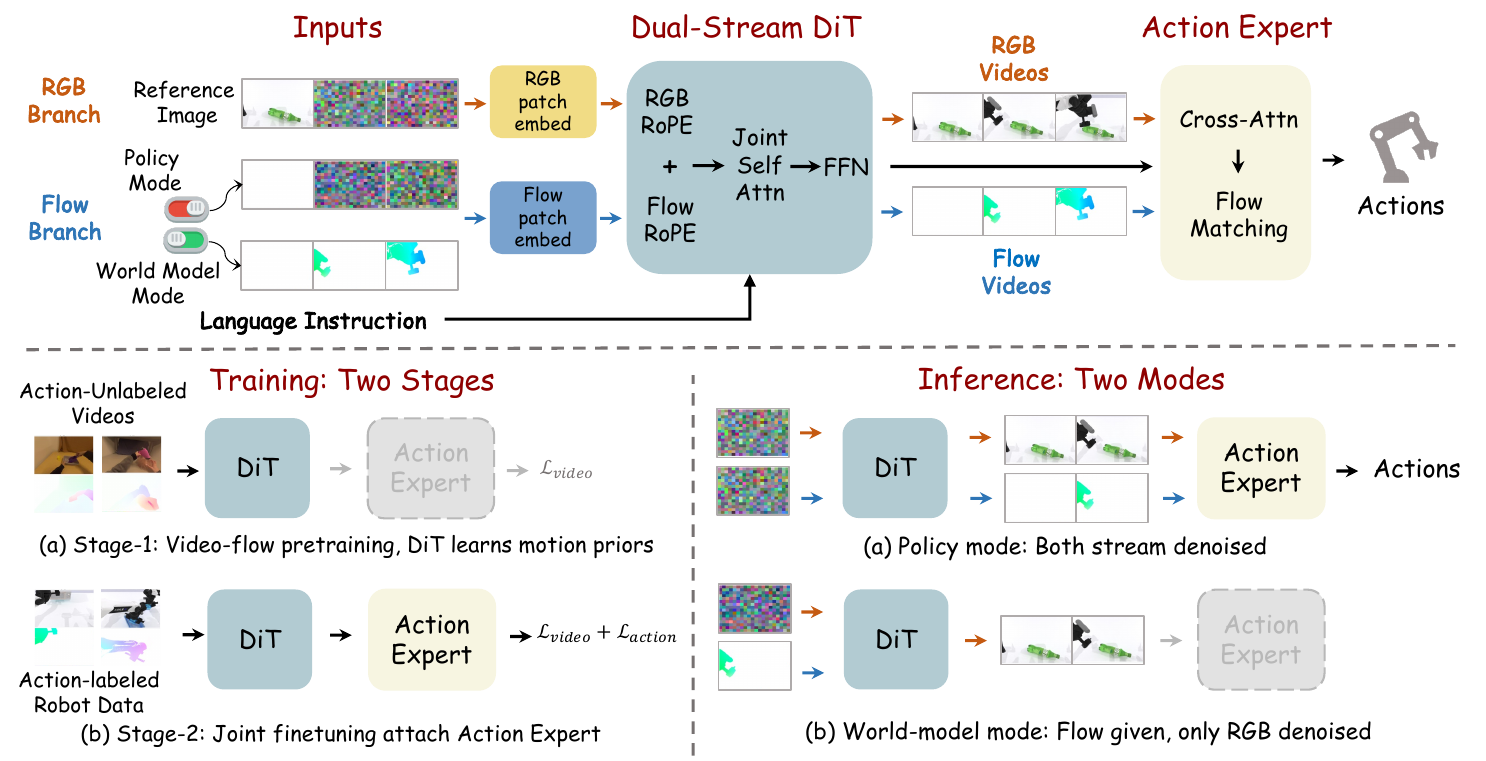}
  \caption{
    \textbf{Overview of FlowWAM.}
    FlowWAM encodes RGB frames and optical flow as two coupled video streams in a shared diffusion transformer.
    In policy mode, the model jointly generates future RGB and flow latents, and an action expert decodes the video generator's intermediate features into executable robot actions.
    In world-model mode, desired flow trajectories are provided as conditioning, and the model generates future RGB frames that follow the specified motion.
  }
  \label{fig:framework}
  \vspace{-0.3cm}
\end{figure*}

\subsection{Overview}
\label{sec:overview}

FlowWAM uses optical flow as the unified action representation that connects pretrained video priors with executable robot control.
Given a reference image $I_0$ and a language instruction $\tau$, it jointly models RGB and flow videos with a dual-stream diffusion transformer.
The same video generator supports two use modes. When flow is generated, FlowWAM acts as a policy whose motion plan is decoded into actions. When flow is provided, FlowWAM acts as a world model that renders an RGB future consistent with the specified motion.
An overview of the full framework is shown in Fig.~\ref{fig:framework}.
We describe the dual-stream generation in Section~\ref{sec:dual_stream}, the action expert in Section~\ref{sec:action_expert}, and the training procedure in Section~\ref{sec:training}.

\subsection{Unified Video-Flow Generation}
\label{sec:dual_stream}

\paragraph{Flow RGB Encoding.}
We first convert action-relevant motion flow into the same image format as RGB video frames.
Given consecutive frames, the optical flow field $\mathbf{f}_t \in \mathbb{R}^{H \times W \times 2}$ records per-pixel displacement $(u, v)$, capturing both where motion occurs and how visible points move.
We convert it into an RGB image with an HSV color-wheel encoding $\phi$:
\begin{equation}
\label{eq:hsv_encoding}
F_t = \phi(\mathbf{f}_t): \quad
\text{H} = \frac{\operatorname{atan2}(v, u) + \pi}{2\pi}, \quad
\text{S} = \frac{\|\mathbf{f}_t\|}{m}, \quad
\text{V} = 1,
\end{equation}
where $m$ is a normalization constant for the flow magnitude and $\text{H}, \text{S}, \text{and } \text{V}$ denote the hue, saturation, and value channels, respectively.
Under the chosen normalization, the encoding is invertible: $\phi^{-1}(F_t)$ recovers the numerical flow field.
Since flow RGB $F_t$ is format-identical to scene frames $I_t$, action-relevant motion can be directly processed by the same VAE encoder and video generator without a separate action tokenizer.

\paragraph{Dual-Stream Architecture.}
RGB frames and flow frames are independently encoded by the same frozen VAE encoder $\mathcal{E}$, producing RGB latents $\mathbf{z} = \mathcal{E}(\mathbf{V})$ and flow latents $\mathbf{z}^f = \mathcal{E}(\mathbf{F})$, where $\mathbf{V}$ and $\mathbf{F}$ denote the RGB and flow video sequences, respectively.
The only stream-specific parameters are the patch embedding layer and output head for each stream, and all transformer blocks are shared.
Inside each self-attention layer, RGB and flow tokens are concatenated for joint attention and then split back into their respective streams, and we apply rotary position embeddings (RoPE) independently to each stream.
This joint attention mechanism enables deep spatiotemporal interaction between the RGB and flow streams, while the shared transformer blocks ensure that the model fully retains the generative capacity of the pretrained video generators.

\paragraph{Two Operating Modes.}
Once RGB and flow are modeled as parallel streams, policy and world-model inference differ only in whether flow is unknown or observed.
In both modes the first latent frame of each stream is fixed to the clean VAE encoding of the reference observation (image-to-video conditioning), and all subsequent latent frames are produced by iterative denoising.

In \textbf{policy mode}, the remaining frames of both streams are initialized from Gaussian noise and jointly denoised. The dual-stream DiT simultaneously synthesizes a future RGB rollout and its corresponding flow video, from which the action expert reads intermediate features to decode executable actions.
In \textbf{world-model mode}, the flow latents $\mathbf{z}^f$ are instead set to the clean VAE encoding of a desired motion trajectory and held fixed throughout sampling, while only the RGB latents are initialized from noise. The model then renders a future video that is consistent with the specified motion, enabling action-conditioned prediction for planning and evaluation.
Thus, flow is the shared action representation in the framework. It is \emph{generated} for policy inference and \emph{provided} for controllable world modeling.

\subsection{Action Prediction from Flow Latents}
\label{sec:action_expert}

FlowWAM first predicts dense RGB and flow latents with the dual-stream DiT, and then translates this motion plan into the robot's low-level action space.
We therefore use an action expert, a transformer that cross-attends to the per-layer flow and RGB hidden states of the dual-stream video generator, conditions on the current proprioceptive state, and predicts an $N$-step action chunk $\hat{\mathbf{a}}_0 \in \mathbb{R}^{N\times d_a}$ under a flow-matching objective.

At training time the action expert can read hidden states computed from clean visual latents (the frozen VAE encoding). At inference it reads hidden states derived from the latents generated by the dual-stream model, which may contain residual denoising error.
To align these distributions, we mix noise into the latents fed to the video generator on a fraction $p{=}0.5$ of training steps and pass the sampled noise level to the action expert as an auxiliary embedding,
\begin{equation}
\label{eq:stochastic_cond}
\tilde{\mathbf{z}}^f = (1-\sigma)\,\mathbf{z}^f + \sigma\,\bm{\epsilon}^f, \quad
\tilde{\mathbf{z}} = (1-\sigma)\,\mathbf{z} + \sigma\,\bm{\epsilon}^r, \quad
\sigma \sim \mathcal{U}[0,1],
\end{equation}
with independent $\bm{\epsilon}^f, \bm{\epsilon}^r \sim \mathcal{N}(\mathbf{0}, \mathbf{I})$. The remaining $1{-}p$ training steps see pure clean latents.

\subsection{Training}
\label{sec:training}

FlowWAM uses one training formulation for both action-labeled robot data and action-unlabeled video data.
With action labels, the dual-stream video generator and action expert are optimized together. Without labels, the same RGB-flow video objective still trains the dual-stream video generator.

\paragraph{Video Generation Loss.}
The dual-stream DiT is trained with a flow matching objective~\citep{lipman2023flow} over both RGB and flow latents.
At each training step, a shared timestep $t \sim \mathcal{U}[0, 1]$ is sampled for both streams.
The noisy latents are constructed as $\mathbf{z}_t = (1-t)\,\mathbf{z}_0 + t\,\bm{\epsilon}$ and $\mathbf{z}^f_t = (1-t)\,\mathbf{z}^f_0 + t\,\bm{\epsilon}'$, and the model predicts the velocity fields $v_\theta(\mathbf{z}_t, t)$ and $v^f_\theta(\mathbf{z}^f_t, t)$.
The video loss is a weighted combination of the two streams:
\begin{equation}
\label{eq:loss_video}
\mathcal{L}_{\text{video}} = (1 - \lambda_f)\,\mathcal{L}_{\text{RGB}} + \lambda_f\,\mathcal{L}_{\text{flow}},
\end{equation}
where $\mathcal{L}_{\text{RGB}}$ and $\mathcal{L}_{\text{flow}}$ are the mean squared errors between predicted and target velocities for the RGB and flow streams, respectively.
To match the image-to-video inference distribution, the first latent frame of each noisy stream is replaced with its clean conditioning latent, and the loss on this frame is correspondingly masked out. We additionally perturb the conditioning frame with a small Gaussian noise to mimic the residual error of the most recent observation in autoregressive rollouts.

Manipulation flow is spatially sparse, with motion concentrated around the embodiment, manipulated objects and contact-relevant regions.
To avoid static-background dominance, we apply \emph{motion-aware reweighting} to $\mathcal{L}_{\text{flow}}$.
Per-location weights are computed from the channel-averaged deviation of the flow latent from the reference frame:
\begin{equation}
\label{eq:motion_weight}
w_{\text{motion}} = 1 + \alpha \cdot \frac{\langle|\mathbf{z}^f - \mathbf{z}^f_{\text{ref}}|\rangle_c}{\max\,\langle|\mathbf{z}^f - \mathbf{z}^f_{\text{ref}}|\rangle_c},
\end{equation}
where $\langle\cdot\rangle_c$ averages over latent channels and $\alpha$ controls the boosting strength, directing learning toward motion-rich regions.

\paragraph{Action Prediction Loss and Total Objective.}
When action labels are available, the action expert is supervised to predict the ground-truth action chunk from the dual-stream DiT's per-layer hidden states under an independent flow-matching schedule, yielding a per-chunk loss $\mathcal{L}_{\text{action}}$.
The labeled-data objective is
\begin{equation}
\label{eq:loss_total}
\mathcal{L} = \mathcal{L}_{\text{video}} + \lambda_a\,\mathcal{L}_{\text{action}},
\end{equation}
where $\lambda_a$ balances visual motion modeling and action decoding.

\paragraph{Action-Free Pretraining on Unlabeled Videos.}
The same formulation also enables action-unlabeled video pretraining.
Because $\mathcal{L}_{\text{video}}$ depends only on RGB frames and extracted optical flow, the dual-stream video generator can be trained on \emph{any} video corpus, including egocentric human-manipulation videos without robot action labels.
We therefore adopt a two-stage recipe: (i)~a \emph{video-only} stage that updates only the dual-stream DiT under $\mathcal{L}_{\text{video}}$, and (ii)~a \emph{joint} stage on robot demonstrations that attaches the action expert and optimizes the full labeled-data objective.
This decouples visual motion modeling from embodiment-specific action decoding, allowing motion priors learned from human video to transfer to robot control without changing the flow representation.

\begin{table*}[t]
    \setlength{\abovecaptionskip}{0pt}
    \setlength{\belowcaptionskip}{7pt}
    \centering
    \caption{\textbf{RoboTwin 2.0 policy success rates under \emph{Clean} and \emph{Random} settings.} PT: pretraining; the average is over all 50 tasks, with the full per-task breakdown in Appendix Table~\ref{tab:robotwin-full}.}
    \label{tab:robotwin-short}
    \footnotesize
    \setlength{\tabcolsep}{3pt}
    \begin{adjustbox}{width=\textwidth,center}
    \begin{tabular}{>{\centering\arraybackslash}m{2.6cm} *{16}{c}}
      \toprule
    \multirow{2}{*}{\textbf{\makecell[c]{Task}}}
      & \multicolumn{2}{c}{$\mathbf{\pi}_{\mathbf{0.5}}$}
      & \multicolumn{2}{c}{\textbf{X-VLA}}
      & \multicolumn{2}{c}{\textbf{Motus}}
      & \multicolumn{2}{c}{\makecell[c]{\textbf{GigaWorld}\\\textbf{Policy}}}
      & \multicolumn{2}{c}{\textbf{X-WAM}}
      & \multicolumn{2}{c}{\textbf{Fast-WAM}}
      & \multicolumn{2}{c}{\makecell[c]{\textbf{FlowWAM}\\\textbf{w/o PT}}}
      & \multicolumn{2}{c}{\makecell[c]{\textbf{FlowWAM}\\\textbf{w/ PT}}} \\
      & \textbf{\emph{Clean}} & \textbf{\emph{Rand.}} & \textbf{\emph{Clean}} & \textbf{\emph{Rand.}} & \textbf{\emph{Clean}} & \textbf{\emph{Rand.}} & \textbf{\emph{Clean}} & \textbf{\emph{Rand.}} & \textbf{\emph{Clean}} & \textbf{\emph{Rand.}} & \textbf{\emph{Clean}} & \textbf{\emph{Rand.}} & \textbf{\emph{Clean}} & \textbf{\emph{Rand.}} & \textbf{\emph{Clean}} & \textbf{\emph{Rand.}} \\
      \midrule
\textit{Adjust Bottle} & 79\% & 83\% & \textbf{100\%} & 99\% & 89\% & 93\% & \textbf{100\%} & \textbf{100\%} & \textbf{100\%} & 99\% & \textbf{100\%} & \textbf{100\%} & 88\% & 98\% & 99\% & \textbf{100\%} \\
\textit{Beat Block Hammer} & 63\% & 50\% & 92\% & 88\% & 95\% & 88\% & 86\% & 86\% & 98\% & 96\% & \textbf{99\%} & 97\% & 75\% & 75\% & \textbf{99\%} & \textbf{100\%} \\
\textit{Grab Roller} & 90\% & 89\% & \textbf{100\%} & \textbf{100\%} & \textbf{100\%} & \textbf{100\%} & \textbf{100\%} & \textbf{100\%} & \textbf{100\%} & \textbf{100\%} & \textbf{100\%} & \textbf{100\%} & \textbf{100\%} & \textbf{100\%} & \textbf{100\%} & 99\% \\
\textit{Handover Mic} & 28\% & 18\% & 0\% & 0\% & 78\% & 63\% & 72\% & 72\% & 88\% & 89\% & \textbf{99\%} & \textbf{100\%} & 89\% & 70\% & \textbf{99\%} & 99\% \\
\textit{Hanging Mug} & 3\% & 3\% & 23\% & 27\% & 38\% & 38\% & 16\% & 12\% & 46\% & 55\% & 58\% & 62\% & 50\% & 50\% & \textbf{65\%} & \textbf{68\%} \\
\textit{Lift Pot} & 0\% & 0\% & 99\% & \textbf{100\%} & 96\% & 99\% & 98\% & 98\% & \textbf{100\%} & 99\% & \textbf{100\%} & \textbf{100\%} & 96\% & 98\% & 98\% & \textbf{100\%} \\
\textit{Move Can Pot} & 29\% & 27\% & 89\% & 86\% & 34\% & 74\% & 76\% & 78\% & 80\% & 84\% & 90\% & 88\% & 62\% & 60\% & \textbf{96\%} & \textbf{97\%} \\
\textit{Move Pillbottle Pad} & 33\% & 29\% & 73\% & 71\% & 93\% & 96\% & 90\% & 90\% & 96\% & 98\% & \textbf{100\%} & \textbf{99\%} & 82\% & 79\% & \textbf{100\%} & 98\% \\
\textit{Move Playingcard Away} & 59\% & 67\% & 93\% & 98\% & \textbf{100\%} & 96\% & 78\% & 72\% & \textbf{100\%} & 98\% & \textbf{100\%} & \textbf{100\%} & 71\% & 71\% & \textbf{100\%} & 99\% \\
\textit{Open Laptop} & 19\% & 35\% & 93\% & \textbf{100\%} & 95\% & 91\% & 96\% & 98\% & 96\% & 97\% & 98\% & \textbf{100\%} & 91\% & 98\% & \textbf{99\%} & 99\% \\
\textit{......(50 tasks)} &&&&&&&&&&&&&&&& \\
\textit{Pick Diverse Bottles} & 5\% & 3\% & 58\% & 36\% & 90\% & 91\% & 82\% & 70\% & \textbf{91\%} & \textbf{92\%} & 80\% & 85\% & 68\% & 68\% & 90\% & 91\% \\
\textit{Pick Dual Bottles} & 10\% & 6\% & 47\% & 36\% & 96\% & 90\% & 86\% & 86\% & 99\% & \textbf{100\%} & \textbf{100\%} & 96\% & 75\% & 91\% & \textbf{100\%} & 99\% \\
\textit{Place A2b Right} & 62\% & 57\% & 36\% & 36\% & 91\% & 87\% & 90\% & 92\% & 92\% & 89\% & 93\% & \textbf{99\%} & 82\% & 86\% & \textbf{99\%} & 96\% \\
\textit{Place Burger Fries} & 66\% & 70\% & 94\% & 94\% & 98\% & 98\% & 98\% & 96\% & 97\% & \textbf{99\%} & 96\% & \textbf{99\%} & \textbf{100\%} & 91\% & 99\% & \textbf{99\%} \\
\textit{Place Container Plate} & 71\% & 78\% & 97\% & 95\% & \textbf{98\%} & 99\% & \textbf{98\%} & 96\% & \textbf{98\%} & \textbf{100\%} & 96\% & \textbf{100\%} & \textbf{98\%} & 97\% & \textbf{98\%} & 98\% \\
\textit{Place Empty Cup} & 75\% & 86\% & \textbf{100\%} & 98\% & 99\% & 98\% & 90\% & 90\% & 98\% & 99\% & \textbf{100\%} & \textbf{100\%} & 92\% & 93\% & \textbf{100\%} & \textbf{100\%} \\
\textit{Place Mouse Pad} & 21\% & 26\% & 70\% & 70\% & 66\% & 68\% & 88\% & 90\% & 84\% & 86\% & 83\% & 89\% & 68\% & 68\% & \textbf{96\%} & \textbf{94\%} \\
\textit{Put Bottles Dustbin} & 12\% & 9\% & 74\% & 77\% & 81\% & 79\% & 72\% & 70\% & 85\% & \textbf{95\%} & 95\% & 90\% & 63\% & 59\% & \textbf{96\%} & 94\% \\
\textit{Scan Object} & 42\% & 38\% & 14\% & 36\% & 67\% & 66\% & 60\% & 64\% & 86\% & 79\% & 89\% & 92\% & 60\% & 50\% & \textbf{92\%} & \textbf{94\%} \\
\textit{Stack Blocks Two} & 48\% & 56\% & 92\% & 87\% & \textbf{100\%} & 98\% & \textbf{100\%} & 94\% & \textbf{100\%} & \textbf{100\%} & \textbf{100\%} & \textbf{100\%} & 95\% & 93\% & \textbf{100\%} & \textbf{100\%} \\
\textit{Stack Bowls Two} & 78\% & 66\% & 96\% & 93\% & 98\% & \textbf{98\%} & 96\% & 92\% & 98\% & \textbf{98\%} & 92\% & \textbf{98\%} & \textbf{100\%} & 91\% & 95\% & 95\% \\
      \midrule
      \textbf{\textit{Average (\%)}} & 42.98 & 43.84 & 72.88 & 72.84 & 88.66 & 87.02 & 86.36 & 85.04 & 89.76 & 90.68 & 91.88 & 91.78 & 82.40 & 80.80 & \textbf{92.94} & \textbf{92.14} \\
      \bottomrule
    \end{tabular}
    \end{adjustbox}
    \vspace{-0.3cm}
  \end{table*}

\section{Experiments}
\label{sec:experiments}

The central claim of FlowWAM is that optical flow can serve as a unified action representation for both policy inference and world modeling.
To assess whether this claim holds in practice, we design our experiments around three concrete properties:
\begin{itemize}
    \item \textbf{Decodability for control:} predicted flow can be decoded by the action expert into reliable robot actions, rather than only matching visual motion patterns.
    \item \textbf{Scalability via action-unlabeled video pretraining:} flow extracted from action-unlabeled egocentric human videos transfers reusable motion priors to downstream policy learning.
    \item \textbf{Reliability for world modeling:} dense flow conditioning supplies a per-pixel motion field that tightly steers video generation to follow the requested action trajectory.
\end{itemize}
We test the first two properties on RoboTwin~2.0 (Section~\ref{sec:exp_policy}), the third on WorldArena (Section~\ref{sec:exp_worldarena}), and additionally verify that the same representation transfers to real-world single-arm and dual-arm robot platforms (Section~\ref{sec:exp_realworld}).
Section~\ref{sec:exp_analysis} then investigates the design choices behind these properties and quantitatively validates decodability via per-task flow error.

\subsection{Manipulation Policy}
\label{sec:exp_policy}

\paragraph{Setup.}
We evaluate manipulation policy on RoboTwin~2.0~\citep{robotwin2}, which provides 50 bimanual tasks under a \emph{Clean} setting with fixed layout and lighting and a \emph{Random} setting with randomized object pose, distractors, lighting, and background.
Models are trained on 50 demonstrations under \emph{Clean} and 500 under \emph{Random} together, and we report success rate over 100 rollouts per task.

We compare against direct-action VLA policies ($\pi_{0.5}$~\citep{pi_0}, X-VLA~\citep{xlva}) and unified world action models that pair video prediction with action decoding (Motus~\citep{motus}, GigaWorld-Policy~\citep{gigaworld-policy}, X-WAM~\citep{guo2026unified}, Fast-WAM~\citep{fast-wam}). To ensure a fair comparison, all WAM baselines are evaluated with their respective pretraining protocols on large-scale video or action-unlabeled data.
FlowWAM is reported with and without action-unlabeled ego-video pretraining on EgoDex~\citep{egodex}.
Table~\ref{tab:robotwin-short} shows a representative subset, with the per-task breakdown in Appendix Table~\ref{tab:robotwin-full}.

\paragraph{Results.}
FlowWAM reaches 92.94\% success on \emph{Clean} and 92.14\% on \emph{Random} in Table~\ref{tab:robotwin-short}, leading both VLA and WAM baselines under both settings.
The advantage is consistent with using flow as a video-native action stream. A dense image-space displacement field gives the action expert a more explicit motion plan than discrete action tokens, and is reliably decoded into joint actions rather than only improving video prediction.

This also supports scalability. Pretraining on action-unlabeled human egocentric videos lifts FlowWAM further, with a clearly larger gain under \emph{Random} settings than under \emph{Clean}. This matches the expectation that randomized scenes benefit most from visual robustness accumulated by the video generator.
The same table also shows that even without ego-video pretraining FlowWAM already surpasses VLA baselines, indicating that the flow stream itself contributes the bulk of the gain and pretraining acts as an amplifier.
This pattern suggests that the policy benefit comes from exposing predicted motion as a dense flow plan before action decoding, rather than relying only on direct low-level action prediction.

\begin{table}[t]
\centering
\caption{\textbf{Main results on WorldArena.}
\textbf{Best} and \underline{second-best} per column are marked.
The \emph{Cond.}\ column lists each method's action-conditioning signal. We prioritize comparisons with models that provide open-source implementations or technical reports to ensure methodological transparency.}
\label{tab:worldarena_main}
\footnotesize
\setlength{\tabcolsep}{3pt}
\renewcommand{\arraystretch}{1.0}
\begin{adjustbox}{width=0.9\textwidth}
\begin{tabular}{l l c c c c c c c}
\toprule
\textbf{Method}
  & \textbf{Cond.}
  & \makecell{\textbf{JEPA}\\\textbf{Sim.} $\uparrow$}
  & \makecell{\textbf{Subj.}\\\textbf{Cons.} $\uparrow$}
  & \makecell{\textbf{Bg.}\\\textbf{Cons.} $\uparrow$}
  & \makecell{\textbf{Traj.}\\\textbf{Acc.} $\uparrow$}
  & \makecell{\textbf{Depth}\\\textbf{Acc.} $\uparrow$}
  & \makecell{\textbf{Sem.}\\\textbf{Align.} $\uparrow$}
  & \textbf{EWMScore} $\uparrow$ \\
\midrule
\rowcolor{gray!15}
\multicolumn{9}{l}{\textit{General video foundation models}} \\
CogVideoX~\citep{cogvideox}                         & Text    & 94.84             & 80.97             & 88.38             & 34.79             & 91.09             & 89.70             & 58.79             \\
Veo 3.1~\citep{veo31}                               & Text    & 57.97             & 75.82             & \textbf{91.67}    & 11.36             & 74.27             & 83.79             & 57.77             \\
Wan 2.6~\citep{wan26}                               & Text    & 72.57             & 73.15             & 84.29             & 12.18             & 75.43             & 88.09             & 59.80             \\
Cosmos-Predict 2.5 (text)~\citep{cosmospredict25}   & Text    & 46.69             & 66.19             & 73.98             & 11.60             & 70.71             & 86.34             & 53.06             \\
ABot-PhysWorld (text)~\citep{abotphysworld}         & Text    & 90.36             & 80.56             & 89.02             & 31.50             & 71.99             & 89.58             & \underline{62.63} \\
\midrule
\rowcolor{gray!15}
\multicolumn{9}{l}{\textit{Action-conditioned robot world models}} \\
Cosmos-Predict 2.5 (action)~\citep{cosmospredict25} & Action  & 90.65             & 67.24             & 72.72             & 27.49             & 90.35             & \underline{89.86} & 54.29             \\
IRASim~\citep{irasim}                               & Action  & 93.72             & 72.82             & 78.69             & 35.92             & 93.50             & 89.43             & 56.15             \\
Ctrl-World~\citep{ctrlworld}                        & Action  & 92.77             & \textbf{83.56}    & \underline{90.30} & 48.20             & 93.25             & 88.68             & 59.98             \\
Vidar~\citep{vidar}                                 & Img-Act & 61.10             & 70.37             & 78.24             & 21.26             & 79.77             & 88.46             & 51.92             \\
Genie Envisioner~\citep{genieenvisioner}            & Img-Act & 33.81             & 74.53             & 87.54             & 2.63              & 86.83             & 85.98             & 41.37             \\
GigaWorld-1~\citep{gigaworld}                       & Img-Act & \underline{96.77} & 80.97             & 86.43             & \underline{54.27} & \underline{98.44} & 89.42             & 62.34             \\
\midrule
\rowcolor{gray!25}
\textbf{FlowWAM (Ours)}                             & \textbf{Flow} & \textbf{97.14} & \underline{82.46} & 89.97 & \textbf{64.26} & \textbf{98.97} & \textbf{89.93} & \textbf{63.71} \\
\bottomrule
\end{tabular}
\end{adjustbox}
\vspace{-0.2cm}
\end{table}

\subsection{Action-Conditioned World Modeling}
\label{sec:exp_worldarena}

\paragraph{Setup.}
We evaluate action-conditioned world modeling on WorldArena~\citep{worldarena}, a unified benchmark that jointly scores video perception quality and action reliability over 121-frame rollouts at 24\,fps.
Each episode provides an initial frame, a language instruction, and the recorded joint-action trajectory, from which the model must synthesize the future video.
We report six representative metrics together with the overall EWMScore, with the full 16-metric breakdown in Appendix Table~\ref{tab:worldarena_full}.

Two families of baselines cover the dominant paradigms.
General video and foundation world models (CogVideoX~\citep{cogvideox}, Veo~3.1~\citep{veo31}, Wan~2.6~\citep{wan26}, Cosmos-Predict~2.5~\citep{cosmospredict25}, ABot-PhysWorld~\citep{abotphysworld}) test whether large-scale visual priors alone suffice for embodied prediction.
Action-conditioned robot world models cover both numerical action conditioning (IRASim~\citep{irasim}, Ctrl-World~\citep{ctrlworld}) and image-space action conditioning (Genie~Envisioner~\citep{genieenvisioner}, GigaWorld-1~\citep{gigaworld}, Vidar~\citep{vidar}), which are designed for manipulation dynamics but do not adopt flow as the conditioning signal.

\paragraph{Results.}
FlowWAM achieves the best overall EWMScore of 63.71 in Table~\ref{tab:worldarena_main}.
The diagnostic axis for reliability is Trajectory Accuracy, which scores how faithfully a generated rollout follows the given action.
Text-conditioned and numerical-action-conditioned baselines remain low in this column, image-space actions yield partial improvements, and FlowWAM gives the strongest result.
This reliability gain does not come at the cost of visual quality. FlowWAM remains close to the best baselines on Subject and Background Consistency, while leading the second-best method by nearly ten points on Trajectory Accuracy.
FlowWAM also tops Depth Accuracy, which is consistent with dense flow providing the generator with a per-pixel motion field to denoise against.

The grouped baselines in Table~\ref{tab:worldarena_main} reveal a representation-level pattern. Text, numerical actions, and image-action cues can preserve appearance, but they leave trajectory control under-specified. Flow conditioning instead provides a dense visual motion field, which explains why FlowWAM's largest margin appears on Trajectory Accuracy rather than appearance-only metrics.

\subsection{Real-World Manipulation}
\label{sec:exp_realworld}

\paragraph{Setup.}
We further evaluate FlowWAM on real-world robot platforms to test whether the flow-based action representation transfers beyond simulated benchmarks.
The suite contains seven tasks across two platforms.
The single-arm Franka group (\textit{Stack Bowls}, \textit{Place in Drawer}, \textit{Put in Plate}, \textit{Place Two Cups}) stresses spatial precision and object contact, while the dual-arm ARX group (\textit{Fold Towel}, \textit{Stack Bowls}, \textit{Clean Plate}) additionally requires bimanual coordination and longer-horizon motion.
For each task, all methods are trained on the same 100 teleoperated demonstrations and evaluated over 10 trials with randomized object poses, with $\pi_{0.5}$~\citep{pi_0} and Motus~\citep{motus} as baselines.

\begin{figure*}[t]
    \centering
    \includegraphics[width=\linewidth]{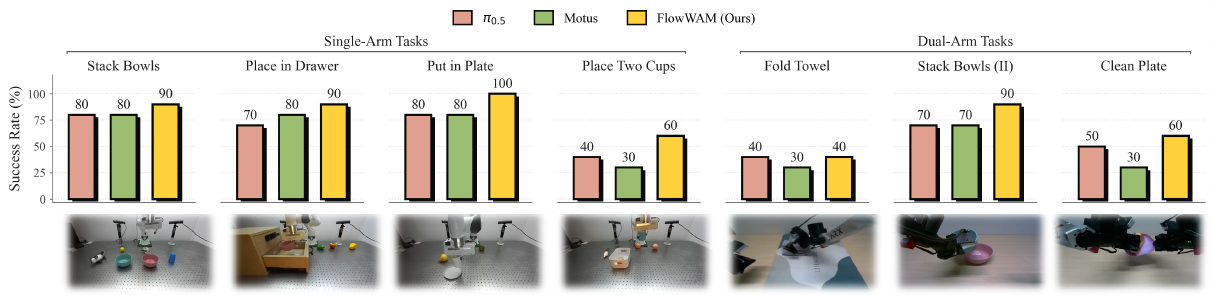}
    \caption{
        \textbf{Real-world success rates across single-arm and dual-arm manipulation tasks.}
    }
    \label{fig:realworld_main}
\end{figure*}

\paragraph{Results.}
Figure~\ref{fig:realworld_main} shows that FlowWAM reaches the highest success rate on every task, averaging 75.7\% across the suite.
The two baselines, $\pi_{0.5}$ and Motus, average 61.4\% and 57.1\%, respectively, and the performance gap relative to FlowWAM is larger on the dual-arm tasks than on the single-arm tasks, where bimanual coordination and longer horizons place a stronger demand on the underlying action representation.
This suggests that flow is especially useful when success depends on coordinating multiple robot parts, since flow represents the coupled arm motion directly in image space rather than only through low-dimensional joint states.
Qualitative rollouts in Appendix~\ref{sec:appx_realworld_policy} (Figure~\ref{fig:appx_realworld_policy}) further visualize how the predicted flow plan aligns with robot trajectories on both platforms.

\subsection{Analysis}
\label{sec:exp_analysis}
\begin{figure}[t]
\centering
\setlength{\abovecaptionskip}{4pt}
\setlength{\belowcaptionskip}{0pt}
\footnotesize
\setlength{\tabcolsep}{4pt}
\renewcommand{\arraystretch}{0.95}
\begin{minipage}[t]{0.34\linewidth}
\centering
{\footnotesize (a) Policy mode (RoboTwin).}\\[2pt]
\begin{tabular}{l c}
\toprule
\textbf{Variant} & \textbf{Succ.\ (\%)} \\
\midrule
Numerical actions               & 69.8 \\
Raw $(u,v)$ flow                & 72.3 \\
w/o flow-loss reweighting       & 83.9 \\
w/o stochastic AE cond.\        & 82.1 \\
\textbf{FlowWAM (full)}         & \textbf{89.8} \\
\bottomrule
\end{tabular}
\end{minipage}
\hfill
\begin{minipage}[t]{0.34\linewidth}
\centering
{\footnotesize (b) World mode (WorldArena).}\\[2pt]
\begin{tabular}{l c}
\toprule
\textbf{Conditioning} & \textbf{EWMScore} \\
\midrule
Text only                       & 49.31 \\
Numerical actions               & 54.18 \\
Flow actions ($u,v$)            & 56.72 \\
Image actions (masks)           & 57.84 \\
\textbf{FlowWAM (full)}         & \textbf{65.23}\makebox[0pt][l]{\footnotemark} \\
\bottomrule
\end{tabular}
\end{minipage}
\hfill
\begin{minipage}[t]{0.30\linewidth}
\centering
{\footnotesize (c) Flow quality vs.\ success.}\\[2pt]
\includegraphics[width=\linewidth]{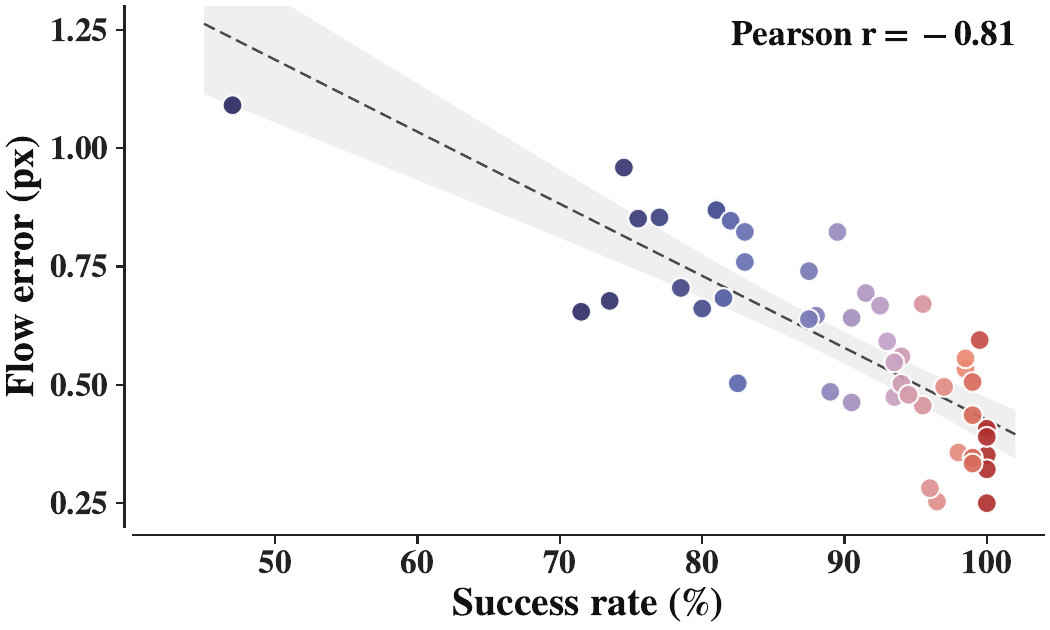}
\end{minipage}
\caption{\textbf{Ablations and flow decodability.}
(a) Policy-side design choices on RoboTwin.
(b) Action-conditioning variants on WorldArena.
(c) Flow error vs.\ policy success on RoboTwin.}
\label{fig:ablation_combined}
\end{figure}
\footnotetext{As official WorldArena evaluation requires remote submission and lacks ground-truth (GT) videos for local scoring, we use a custom validation split for ablations, resulting in a discrepancy with results in Table~\ref{tab:worldarena_main}.}

\paragraph{Ablation Study.}
Panels (a) and (b) of Figure~\ref{fig:ablation_combined} ablate the design choices on both sides of the model.
On the policy side (panel a), replacing flow with numerical action vectors collapses success to 69.8\%, and substituting RGB flow with raw $(u,v)$ tensors breaks the input format expected by the pretrained VAE and DiT (Eq.~\ref{eq:hsv_encoding}). Both results confirm that the flow channel itself, not generic visual conditioning, closes the action-video gap.
Removing motion-aware reweighting (Eq.~\ref{eq:motion_weight}) lets the static background dominate $\mathcal{L}_{\text{flow}}$, while training the action expert without stochastic latent conditioning (Eq.~\ref{eq:stochastic_cond}) leaves it brittle to the residual denoising error of generated latents at inference.

On the world-model side (panel b), holding the video generator fixed and varying only the conditioning input shows the same pattern as our motivation: text-only conditioning leaves motion under-specified, numerical actions provide robot states outside the visual space, raw $(u,v)$ tensors carry dense displacement but not in the image format expected by the pretrained video generator, and image-space masks remain largely static spatial cues that indicate where actions occur without encoding cross-frame motion direction or magnitude.
Only the full RGB flow combines per-pixel motion with a video-native representation, raising EWMScore from 49.31 to 65.23.
The two panels together indicate that flow is required at both ends of the model: as a decodable target on the action side and as a spatially aligned conditioning signal on the generation side.

\paragraph{Flow decodability.}
A central premise of FlowWAM is decodability: predicted flow should match image-space motion and provide the information needed to recover correct robot actions.
We test this by correlating flow prediction quality with task success, rather than attributing all policy gains to the downstream decoder.
Panel (c) of Figure~\ref{fig:ablation_combined} analyzes all 50 RoboTwin tasks.
For each task, we plot the error between FlowWAM's predicted flow and the RAFT-extracted pseudo-ground-truth flow (computed on a newly generated set of 50 demonstrations per task) against the corresponding policy success rate.
Tasks with lower predicted-flow error consistently attain higher success, with a Pearson correlation of $r=-0.81$. Policy gains thus track improvements in flow prediction itself rather than an unrelated decoder shortcut, which provides direct quantitative evidence for decodability.

\begin{figure*}[t]
    \centering
    \includegraphics[width=\linewidth]{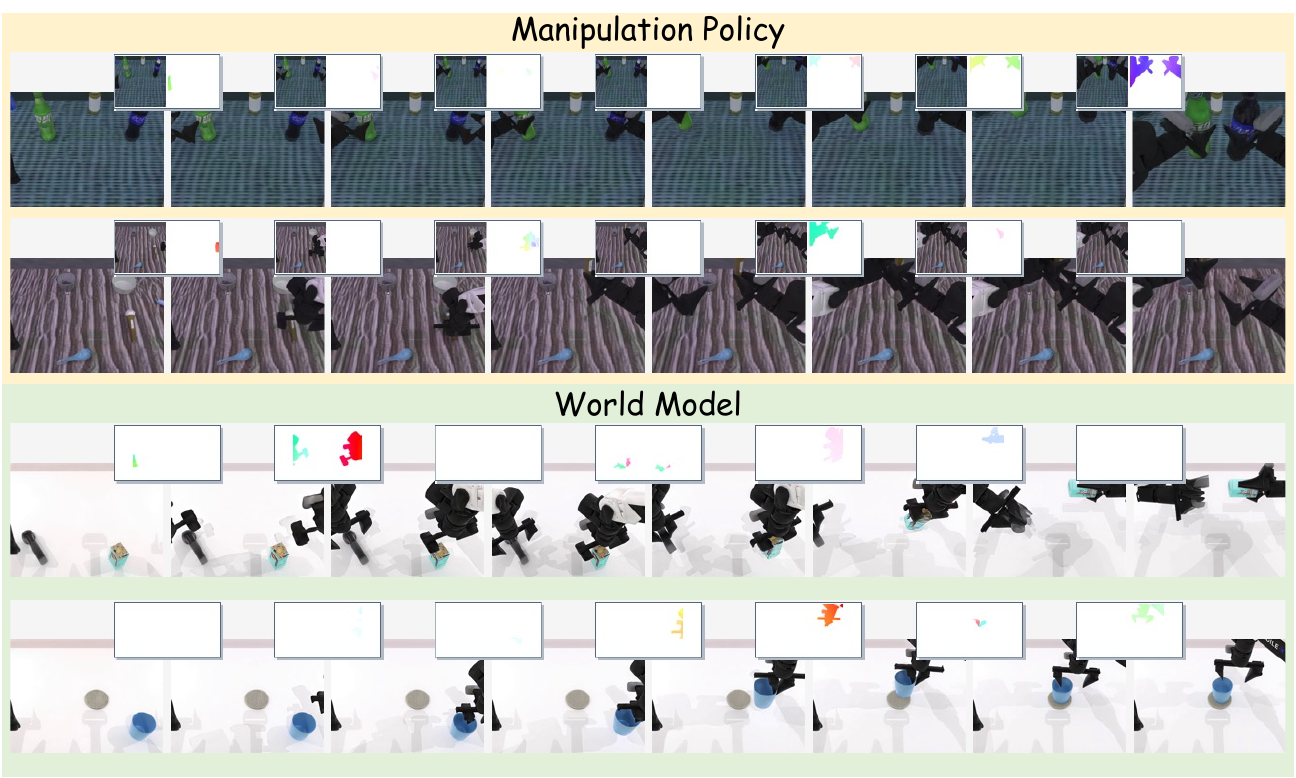}
    \caption{
        \textbf{Qualitative visualization of FlowWAM in policy and world-model modes.}
    }
    \label{fig:qualitative_main}
    \vspace{-0.5cm}
\end{figure*}

\paragraph{Qualitative Results.}
Figure~\ref{fig:qualitative_main} visualizes FlowWAM in its two operating modes.
In policy mode, the predicted flow concentrates on the moving embodiment along a coherent goal-directed trajectory, while the RGB stream preserves the surrounding object context. Decoding this robot-motion plan yields successful task completion across randomized object placements and backgrounds.
In world-model mode, the generated rollout follows the motion prescribed by the conditioning flow, while object identity and the static background remain stable across frames.
Additional rollouts on more RoboTwin tasks under \emph{Clean} and \emph{Random} scenes (Figs.~\ref{fig:appx_robotwin_clean} and~\ref{fig:appx_robotwin_random}) and further world-model examples (Fig.~\ref{fig:appx_world_model_vis}) are provided in Appendix~\ref{sec:appx_visualization}.
\section{Conclusion}
\label{sec:conclusion}

In this paper, we introduced FlowWAM, which establishes optical flow as a unified action representation for WAMs. By encoding robot motion as dense flow videos, we bridge the modality gap between executable control and pretrained video priors. This unified representation allows the same model to generate flow for policy decoding, condition on flow for world modeling, and learn from action-unlabeled egocentric videos. Experiments show that this unified action representation improves both control and world modeling because it preserves dense cross-frame motion while remaining decodable into robot actions. Future work includes scaling action-free pretraining to internet-scale datasets and extending flow-based planning to longer temporal horizons.

\bibliographystyle{unsrtnat}
{\small
\bibliography{references}
}

\newpage
\appendix



\section{Implementation Details and Training Pipeline}
\label{sec:appx_implementation}

This section provides comprehensive implementation details to facilitate the reproduction of the training procedure described in Section~\ref{sec:training}.
We detail the architectural components of FlowWAM (Sec.~\ref{sec:appx_modules}), the two-stage training strategy (Sec.~\ref{sec:appx_two_stage}), dataset construction and preprocessing (Sec.~\ref{sec:appx_data_preprocess}), the multi-view input configuration (Sec.~\ref{sec:appx_inputs}), and the hyperparameter configurations for both stages (Sec.~\ref{sec:appx_hyperparams}).

\subsection{Architectural Components}
\label{sec:appx_modules}

FlowWAM is constructed by integrating motion-specific modules into a pretrained video generator.
The base architecture is Wan2.2-TI2V-5B~\citep{wan22}, a 5-billion parameter image-to-video diffusion transformer (DiT) utilizing a UMT5-XXL text encoder and the Wan2.2 causal VAE.
The pretrained DiT, VAE, and text encoder architectures are preserved. During training, the VAE and text encoder remain frozen, while the DiT parameters are updated according to the stage-specific protocols detailed below.

The dual-stream generator incorporates a dedicated pathway for HSV-encoded flow while maintaining the capacity of the original video generator.
The flow pathway comprises a separate patch embedding layer and output head, both initialized from the corresponding RGB layers. A learnable flow-token embedding is introduced post-patchification to provide a stream-specific identity signal during joint self-attention.
This architecture enables RGB and flow tokens to share transformer blocks and latent geometry while maintaining distinct stream identities.
Since flow frames are encoded using the same VAE as RGB frames, initializing the flow stream with RGB weights provides a compatible image-latent prior, accelerating convergence compared to random initialization.

The action expert is designed to map visual-motion latents to low-level robotic actions.
It is implemented as an AdaLN diffusion transformer with approximately 780M parameters, consisting of 30 layers (matching the depth of the Wan2.2 video DiT) with a hidden dimension of 1024, 16 attention heads, and an FFN dimension of 4096.
Each block integrates self-attention over action tokens, cross-attention to concatenated RGB and flow visual features, and cross-attention to T5 instruction tokens.
Rather than patchifying the VAE latents, these visual features are the per-layer hidden states of the dual-stream video DiT: each action-expert block cross-attends to the concatenated RGB and flow features read out from its paired video-DiT layer.
A proprioceptive token, derived from the normalized 14D joint positions (qpos), is projected to the instruction-embedding dimension and appended to the T5 instruction context, conditioning the predicted action chunk on both the generated motion plan and the robot's initial state.
The prediction head directly predicts the flow-matching velocity of the action chunk, which is integrated by the flow-matching sampler at inference.

\subsection{Two-Stage Training Strategy}
\label{sec:appx_two_stage}

The training pipeline is divided into two stages to decouple general motion representation learning from embodiment-specific action decoding.
Both stages utilize the RGB-flow representation and a flow-matching objective, though they differ in supervision, data sources, and trainable parameters.

\paragraph{Stage 1: Action-Free Motion Pretraining.}
Stage~1 involves pretraining the dual-stream video generator on EgoDex~\citep{egodex}, a dataset of egocentric human manipulation videos without robotic action labels.
This stage focuses on learning manipulation-centric motion priors using RGB frames and extracted optical flow.
Only the video loss $\mathcal{L}_{\text{video}}$ is optimized, with a learning rate of $5\!\times\!10^{-5}$ and updates restricted to the dual-stream DiT.

\paragraph{Stage 2: Joint Embodied Training.}
Stage~2 initializes the video generator from the Stage~1 checkpoint, integrates the action expert, and optimizes the joint objective (Eq.~\ref{eq:loss_total}) on RoboTwin~2.0.
Input streams from the head, left-wrist, and right-wrist cameras are tiled into a $320\times384$ configuration (see Sec.~\ref{sec:appx_inputs}). Each chunk generates a rollout of $T_{\text{pixel}}=9$ pixel frames, from which a 32-step action chunk is decoded (temporal stride 4 between pixel frames).
Training is conducted with a learning rate of $1\!\times\!10^{-4}$, using loss weights $\lambda_f=0.1$ and $\lambda_a=1.0$, and a motion-aware reweighting strength $\alpha=2.0$.
The trainable parameters include the full DiT, the flow stream, and the action expert.

\subsection{Dataset Construction and Preprocessing}
\label{sec:appx_data_preprocess}

\paragraph{RoboTwin Demonstrations.}
For embodied training, we use RoboTwin~2.0 across 50 bimanual manipulation tasks.
Each task contains a \emph{Clean} split with 50 demonstrations under canonical scene configurations and a \emph{Random} split with 500 demonstrations under randomized object poses, distractors, lighting, and backgrounds.
Both splits are used during Stage~2 training so that the action expert observes canonical manipulation trajectories as well as visually diverse scene layouts.
At evaluation time, success rates are computed over 100 rollouts per task under each setting, following the benchmark protocol.

\paragraph{EgoDex Clip Sampling.}
For action-free pretraining, we use EgoDex as an unlabeled egocentric manipulation video source.
Since EgoDex does not provide robot actions, it is used only for the RGB-flow video objective.
We do not apply additional task-level filtering.
Videos are temporally subsampled from 30\,fps to 15\,fps and resized to $320\times256$ to match the RoboTwin head-camera aspect ratio.
During training, clips are sampled from frame buckets $\{17,33,49,65,81\}$, allowing the model to see both short local motions and longer manipulation trajectories.

\paragraph{Optical-Flow Extraction.}
We extract optical flow using RAFT~\citep{raft} for both EgoDex and RoboTwin preprocessing.
For EgoDex, flow is computed directly between adjacent video frames.
For RoboTwin, we construct robot-only flow targets by replaying the recorded robot joint actions in SAPIEN~\citep{sapien} using the robot URDF and rendering the moving embodiment against a static background.
RAFT is then applied to adjacent robot-only frames.
This removes background, lighting, and object-motion artifacts from the flow target, yielding cleaner supervision for robot-induced motion.

\paragraph{HSV Encoding and Magnitude Normalization.}
The numerical flow field is converted into an HSV image using Eq.~\ref{eq:hsv_encoding}, where hue represents direction and saturation represents magnitude.
We use a 25\,px magnitude cap before normalization to prevent large displacements from saturating the color encoding.
Displacements below 0.5\,px are thresholded to zero to suppress flow noise in nearly static regions.

\paragraph{Multi-View Flow Targets.}
For Stage~2 RoboTwin training, RGB observations from the head, left-wrist, and right-wrist cameras are tiled into the T-shape input layout.
Flow supervision is provided for the head-camera region, where the robot-only replay gives a stable global view of embodiment motion.
The wrist-flow regions are filled with a constant placeholder because wrist-view robot-only flow is not available in our preprocessing pipeline.
This keeps the tiled flow input format aligned with the RGB input while avoiding noisy or inconsistent pseudo-labels for wrist cameras.

\subsection{Input Representation and Flow Encoding}
\label{sec:appx_inputs}

\paragraph{Multi-View T-Shape Tiling.}
To process multi-view inputs without specialized encoders, we tile the head and wrist camera views into a single frame.
Following the \emph{T-shape} layout, the head view occupies the top region at its native resolution ($320\times256$), while the left and right wrist views are downsampled and concatenated below it.
The resulting $320\times384$ tile is compatible with the frozen VAE and video generator patch embedding, maintaining architectural consistency across single-view and multi-view scenarios.

\paragraph{Flow Field Generation.}
The flow stream is designed to capture robot-induced motion.
The extraction pipeline follows Sec.~\ref{sec:appx_data_preprocess}: EgoDex flow is computed from adjacent RGB frames, while RoboTwin flow is computed from robot-only replay videos to isolate embodiment motion.
For the multi-view tile, the head-camera flow occupies the head region and the wrist regions use the constant placeholder described above.

\paragraph{HSV Encoding and Normalization.}
The numerical flow field is converted to an HSV image (Eq.~\ref{eq:hsv_encoding}), mapping direction to hue and magnitude to saturation.
We employ the 25\,px magnitude cap described in Sec.~\ref{sec:appx_data_preprocess}.

\subsection{Hyperparameter Summary}
\label{sec:appx_hyperparams}

Table~\ref{tab:appx_hyperparams} summarizes the architectural, data, optimization, and numerical settings used in the two training stages.
All training is performed on 32 NVIDIA H100 GPUs across four nodes.

\begin{table}[!ht]
  \centering
  \small
  \caption{Hyperparameters for the two FlowWAM training stages.}
  \label{tab:appx_hyperparams}
  \begin{adjustbox}{width=0.8\textwidth}
  \renewcommand{\arraystretch}{1.05}
  \begin{tabular}{l l}
    \toprule
    \textbf{Hyperparameter} & \textbf{Value} \\
    \midrule
    \rowcolor{gray!15}
    \multicolumn{2}{l}{\textit{Video Generator}} \\
    \quad Base model                                & Wan2.2-TI2V-5B \\
    \quad VAE / Text encoder                        & Wan2.2 VAE / UMT5-XXL (frozen, bf16) \\
    \quad Flow stream init                          & deep copy of patch embedding \& head \\
    \rowcolor{gray!15}
    \multicolumn{2}{l}{\textit{Action Expert}} \\
    \quad Layers / hidden / heads / FFN             & 30 / 1024 / 16 / 4096 \\
    \quad Parameters                                & $\sim$780M \\
    \quad Action dim                                & 14 \\
    \rowcolor{gray!15}
    \multicolumn{2}{l}{\textit{Inputs (RoboTwin, Stage 2)}} \\
    \quad Per-camera resolution                     & $320 \times 256$ \\
    \quad T-shape tiled resolution                  & $320 \times 384$ \\
    \quad Pixel frames / action steps               & 9 / 32 \\
    \rowcolor{gray!15}
    \multicolumn{2}{l}{\textit{Inputs (EgoDex, Stage 1)}} \\
    \quad Resolution                                & $320 \times 256$ \\
    \quad Frame buckets                             & $\{17,33,49,65,81\}$ \\
    \quad Effective fps                             & 15 (stride 2 from 30\,fps) \\
    \rowcolor{gray!15}
    \multicolumn{2}{l}{\textit{Flow}} \\
    \quad Estimator                                 & RAFT \\
    \quad Magnitude cap                             & 25\,px \\
    \quad Noise threshold                       & 0.5\,px \\
    \rowcolor{gray!15}
    \multicolumn{2}{l}{\textit{Optimization}} \\
    \quad Optimizer                                 & AdamW (weight decay $10^{-2}$) \\
    \quad Learning rate (Stage 1 / Stage 2)         & $5\!\times\!10^{-5}$ / $1\!\times\!10^{-4}$ \\
    \quad Per-GPU batch size (Stage 1 / Stage 2)    & 1 / 16 \\
    \rowcolor{gray!15}
    \multicolumn{2}{l}{\textit{Losses \& schedules}} \\
    \quad Flow weight $\lambda_f$                   & 0.1 \\
    \quad Action weight $\lambda_a$ (Stage 2)       & 1.0 \\
    \quad Motion-boost $\alpha$ (Stage 2)           & 2.0 \\
    \quad Ref-aug strength                          & 0.1 \\
    \rowcolor{gray!15}
    \multicolumn{2}{l}{\textit{Inference (RoboTwin eval)}} \\
    \quad Video / action denoising steps            & 25 / 50 \\
    \quad Execute-and-replan window                 & 25 \\
    \bottomrule
  \end{tabular}
  \end{adjustbox}
\end{table}

\section{Baseline Details}
\label{sec:appx_baselines}

This section provides technical descriptions of the baseline models evaluated in our simulated manipulation and world-modeling experiments. 
For real-world evaluations, we benchmark against $\pi_{0.5}$ and Motus, following the experimental protocol detailed in Section~\ref{sec:exp_realworld} and Appendix~\ref{sec:appx_realworld_setup}.

\subsection{Manipulation Policy Baselines}
\label{sec:appx_policy_baselines}

\begin{itemize}[leftmargin=2em, labelsep=0.45em, itemsep=1pt, topsep=2pt, parsep=0pt, partopsep=0pt]
    \item \textbf{$\pi_{0.5}$}~\citep{pi_0} is a vision-language-action (VLA) foundation model that inherits internet-scale semantic knowledge from a pretrained VLM backbone. It employs a flow-matching action expert to generate continuous action chunks, enabling high-frequency (50\,Hz) control for complex manipulation tasks. We include it as a strong direct-action policy baseline: it predicts low-level actions from visual-language observations without explicitly generating an intermediate video-native motion plan.
    \item \textbf{X-VLA}~\citep{xlva} addresses cross-embodiment heterogeneity through a soft-prompt mechanism. It introduces learnable embeddings for distinct hardware configurations within a flow-matching Transformer architecture, facilitating stable pretraining on diverse, multi-robot datasets. This baseline tests whether embodiment-aware prompting and direct action prediction can match the benefits of representing actions as dense image-space motion.
    \item \textbf{Motus}~\citep{motus} proposes a unified latent-action world model utilizing a Mixture-of-Transformer (MoT) architecture. It leverages optical-flow-derived latent actions to bridge visual dynamics with control signals, enabling the action expert to be pretrained on large-scale unlabeled video data. Motus is the closest policy-side comparison to FlowWAM because both exploit unlabeled video and motion cues, but Motus represents actions in a learned latent-action space rather than generating an explicit optical-flow video stream.
    \item \textbf{GigaWorld-Policy}~\citep{gigaworld-policy} is an action-centered world-action model that learns joint pixel--action dynamics. By utilizing a causal design that prevents future-video tokens from influencing action tokens, it enables efficient action decoding at inference while maintaining auxiliary video-generation supervision during training. We include it to compare against a WAM-style policy that uses video prediction as auxiliary supervision but does not expose dense flow as the shared interface between prediction and action decoding.
    \item \textbf{X-WAM}~\citep{guo2026unified} is a unified 4D world-action model that jointly predicts future multi-view RGB-D videos and robot actions from a pretrained video prior. It augments the video generator with a lightweight depth-adaptation branch for spatial reconstruction and adopts an asynchronous noise schedule that decodes actions in few denoising steps while allocating the full step budget to video. We include it as a strong spatially-aware WAM baseline that conditions on numerically-projected action/state latents rather than exposing motion as a dense image-space flow stream.
    \item \textbf{Fast-WAM}~\citep{fast-wam} is a world-action model that retains video co-training during training but skips explicit future imagination at test time, decoding actions directly for real-time control. It is a close comparison for isolating the role of the video prior: like FlowWAM it benefits from video modeling, but it represents actions with numerical tokens and forgoes an explicit generated motion plan, whereas FlowWAM decodes actions from a jointly generated flow stream.
\end{itemize}

\subsection{World-Modeling Baselines}
\label{sec:appx_worldmodel_baselines}

\paragraph{General Video and Foundation World Models.}
These baselines evaluate the capacity of large-scale generative priors to model embodied dynamics without explicit action conditioning or with only high-level text guidance.
\begin{itemize}[leftmargin=2em, labelsep=0.45em, itemsep=1pt, topsep=2pt, parsep=0pt, partopsep=0pt]
    \item \textbf{CogVideoX}~\citep{cogvideox} is a diffusion Transformer-based video generator that utilizes a 3D causal VAE and an expert Transformer architecture with adaptive LayerNorm to produce temporally consistent, high-resolution videos. It represents the open video-generation baseline family, where embodied rollouts are generated from visual and text context without explicit low-level robot-action conditioning.
    \item \textbf{Veo~3.1}~\citep{veo31} is a state-of-the-art video generator that emphasizes granular narrative control and enhanced realism. In our experiments, it serves as a high-fidelity text-to-video baseline. Its inclusion probes whether very strong generic video priors can compensate for the absence of dense action-conditioned motion guidance in robot manipulation scenes.
    \item \textbf{Wan~2.6}~\citep{wan26} is a multimodal video foundation model designed for cinematic storytelling, supporting multi-shot narratives and precise audiovisual synchronization. Since FlowWAM builds on the Wan video-generation family, this baseline helps separate the contribution of the pretrained video prior from the additional flow-conditioned action representation.
    \item \textbf{Cosmos-Predict~2.5}~\citep{cosmospredict25} is an NVIDIA world foundation model for physical AI. It unifies text-, image-, and video-to-world generation within a single flow-based framework, supporting multi-view outputs and long-horizon simulation. We evaluate it as a broad physical-world prior that is designed for embodied prediction, while still differing from FlowWAM in how low-level manipulation actions are represented and injected.
    \item \textbf{ABot-PhysWorld}~\citep{abotphysworld} is a 14B diffusion Transformer model that incorporates physics-aware annotations and DPO-based post-training to suppress unphysical behaviors (e.g., object penetration) in generated manipulation sequences. It is included as a strong physics-aware foundation baseline, testing whether post-training for physical plausibility alone is sufficient for trajectory-accurate action following.
\end{itemize}

\paragraph{Action-Conditioned Robot World Models.}
These models explicitly incorporate low-level robot actions or image-space action cues as conditioning signals for future prediction.
\begin{itemize}[leftmargin=2em, labelsep=0.45em, itemsep=1pt, topsep=2pt, parsep=0pt, partopsep=0pt]
    \item \textbf{IRASim}~\citep{irasim} is a trajectory-to-video world model that introduces a frame-level action-conditioning module within each Transformer block to strengthen the alignment between predicted frames and fine-grained robot--object interactions. It provides a direct comparison to numerical-action conditioning, where robot state/action vectors are injected into the video generator rather than first converted into image-space motion.
    \item \textbf{Ctrl-World}~\citep{ctrlworld} is a controllable multi-view world model designed for policy-in-the-loop rollouts. It maintains long-horizon consistency through a pose-conditioned memory retrieval mechanism and frame-level action conditioning. This baseline tests whether explicit memory and pose conditioning can provide stability comparable to dense flow conditioning over long manipulation rollouts.
    \item \textbf{Vidar}~\citep{vidar} is a two-stage bimanual manipulation framework that couples a general video diffusion model with a masked inverse dynamics model (MIDM) to extract action-relevant information from generated trajectories. It is relevant to our setting because it also links video generation with bimanual action reasoning, but it relies on a downstream inverse-dynamics module rather than using flow as the conditioning signal for future prediction.
    \item \textbf{Genie Envisioner}~\citep{genieenvisioner} is a unified platform that integrates instruction-conditioned video generation, neural simulation, and policy learning. It employs a lightweight parallel flow-matching decoder to map visual latents to motor commands. We compare against it as a representative end-to-end embodied world-modeling system that couples visual prediction and control without explicitly enforcing optical flow as the shared action interface.
    \item \textbf{GigaWorld-1}~\citep{gigaworld} functions as a world-model data engine. It provides a strong comparison for data-driven stable world modeling, where stability is obtained through scene-level reconstruction and synthetic data generation rather than through dense per-pixel flow trajectories.
\end{itemize}

\section{Additional RoboTwin Results}
\label{sec:appx_robotwin}

Table~\ref{tab:robotwin-full} reports the full per-task RoboTwin~2.0 policy success rates corresponding to the compact main-text comparison in Table~\ref{tab:robotwin-short}.

\begin{table*}[ht]
  \setlength{\abovecaptionskip}{0pt}
  \setlength{\belowcaptionskip}{7pt}
  \centering
  \caption{\textbf{Full RoboTwin 2.0 per-task policy success rates under \emph{Clean} and \emph{Random} scenes.}
  Bolded numbers indicate the best result in each row. FlowWAM is reported with and without large-scale pretraining (PT).}
  \label{tab:robotwin-full}
  \footnotesize
  \setlength{\tabcolsep}{3pt}
  \begin{adjustbox}{width=\textwidth,center}
  \begin{tabular}{>{\centering\arraybackslash}m{2.6cm} *{16}{c}}
    \toprule
    \textbf{\makecell[c]{Simulation Task}}
      & \multicolumn{2}{c}{$\mathbf{\pi}_{\mathbf{0.5}}$}
      & \multicolumn{2}{c}{\textbf{X-VLA}}
      & \multicolumn{2}{c}{\textbf{Motus}}
      & \multicolumn{2}{c}{\makecell[c]{\textbf{GigaWorld}\\\textbf{Policy}}}
      & \multicolumn{2}{c}{\textbf{X-WAM}}
      & \multicolumn{2}{c}{\textbf{Fast-WAM}}
      & \multicolumn{2}{c}{\makecell[c]{\textbf{FlowWAM}\\\textbf{w/o PT}}}
      & \multicolumn{2}{c}{\makecell[c]{\textbf{FlowWAM}\\\textbf{w/ PT}}} \\
      & \textbf{\emph{Clean}} & \textbf{\emph{Rand.}} & \textbf{\emph{Clean}} & \textbf{\emph{Rand.}} & \textbf{\emph{Clean}} & \textbf{\emph{Rand.}} & \textbf{\emph{Clean}} & \textbf{\emph{Rand.}} & \textbf{\emph{Clean}} & \textbf{\emph{Rand.}} & \textbf{\emph{Clean}} & \textbf{\emph{Rand.}} & \textbf{\emph{Clean}} & \textbf{\emph{Rand.}} & \textbf{\emph{Clean}} & \textbf{\emph{Rand.}} \\
    \midrule
\textit{Adjust Bottle} & 79\% & 83\% & \textbf{100\%} & 99\% & 89\% & 93\% & \textbf{100\%} & \textbf{100\%} & \textbf{100\%} & 99\% & \textbf{100\%} & \textbf{100\%} & 88\% & 98\% & 99\% & \textbf{100\%} \\
\textit{Beat Block Hammer} & 63\% & 50\% & 92\% & 88\% & 95\% & 88\% & 86\% & 86\% & 98\% & 96\% & \textbf{99\%} & 97\% & 75\% & 75\% & \textbf{99\%} & \textbf{100\%} \\
\textit{Blocks Ranking Rgb} & 43\% & 35\% & 83\% & 83\% & 99\% & 97\% & 92\% & 96\% & 99\% & 95\% & \textbf{100\%} & \textbf{100\%} & 91\% & 99\% & 99\% & \textbf{100\%} \\
\textit{Blocks Ranking Size} & 8\% & 14\% & 67\% & 74\% & 75\% & 63\% & 44\% & 48\% & 76\% & 82\% & \textbf{94\%} & \textbf{98\%} & 68\% & 65\% & 82\% & 94\% \\
\textit{Click Alarmclock} & 97\% & 93\% & 99\% & 99\% & \textbf{100\%} & \textbf{100\%} & \textbf{100\%} & \textbf{100\%} & 98\% & 99\% & \textbf{100\%} & \textbf{100\%} & \textbf{100\%} & 99\% & 98\% & \textbf{100\%} \\
\textit{Click Bell} & 75\% & 76\% & \textbf{100\%} & \textbf{100\%} & \textbf{100\%} & \textbf{100\%} & \textbf{100\%} & \textbf{100\%} & \textbf{100\%} & \textbf{100\%} & \textbf{100\%} & \textbf{100\%} & \textbf{100\%} & \textbf{100\%} & 95\% & 85\% \\
\textit{Dump Bin Bigbin} & 30\% & 42\% & 79\% & 77\% & 95\% & 91\% & 92\% & \textbf{100\%} & 90\% & 96\% & \textbf{97\%} & 96\% & 89\% & 96\% & 96\% & 98\% \\
\textit{Grab Roller} & 90\% & 89\% & \textbf{100\%} & \textbf{100\%} & \textbf{100\%} & \textbf{100\%} & \textbf{100\%} & \textbf{100\%} & \textbf{100\%} & \textbf{100\%} & \textbf{100\%} & \textbf{100\%} & \textbf{100\%} & \textbf{100\%} & \textbf{100\%} & 99\% \\
\textit{Handover Block} & 18\% & 19\% & 73\% & 37\% & 86\% & 73\% & 80\% & 80\% & 88\% & 79\% & 95\% & \textbf{81\%} & 67\% & 52\% & \textbf{96\%} & 78\% \\
\textit{Handover Mic} & 28\% & 18\% & 0\% & 0\% & 78\% & 63\% & 72\% & 72\% & 88\% & 89\% & \textbf{99\%} & \textbf{100\%} & 89\% & 70\% & \textbf{99\%} & 99\% \\
\textit{Hanging Mug} & 3\% & 3\% & 23\% & 27\% & 38\% & 38\% & 16\% & 12\% & 46\% & 55\% & 58\% & 62\% & 50\% & 50\% & \textbf{65\%} & \textbf{68\%} \\
\textit{Lift Pot} & 0\% & 0\% & 99\% & \textbf{100\%} & 96\% & 99\% & 98\% & 98\% & \textbf{100\%} & 99\% & \textbf{100\%} & \textbf{100\%} & 96\% & 98\% & 98\% & \textbf{100\%} \\
\textit{Move Can Pot} & 29\% & 27\% & 89\% & 86\% & 34\% & 74\% & 76\% & 78\% & 80\% & 84\% & 90\% & 88\% & 62\% & 60\% & \textbf{96\%} & \textbf{97\%} \\
\textit{Move Pillbottle Pad} & 33\% & 29\% & 73\% & 71\% & 93\% & 96\% & 90\% & 90\% & 96\% & 98\% & \textbf{100\%} & \textbf{99\%} & 82\% & 79\% & \textbf{100\%} & 98\% \\
\textit{Move Playingcard Away} & 59\% & 67\% & 93\% & 98\% & \textbf{100\%} & 96\% & 78\% & 72\% & \textbf{100\%} & 98\% & \textbf{100\%} & \textbf{100\%} & 71\% & 71\% & \textbf{100\%} & 99\% \\
\textit{Move Stapler Pad} & 16\% & 18\% & 78\% & 73\% & 83\% & \textbf{85\%} & \textbf{92\%} & 82\% & 67\% & 70\% & 77\% & 64\% & 60\% & 55\% & 89\% & 83\% \\
\textit{Open Laptop} & 19\% & 35\% & 93\% & \textbf{100\%} & 95\% & 91\% & 96\% & 98\% & 96\% & 97\% & 98\% & \textbf{100\%} & 91\% & 98\% & \textbf{99\%} & 99\% \\
\textit{Open Microwave} & 35\% & 37\% & 79\% & 71\% & \textbf{95\%} & 91\% & 74\% & 66\% & 89\% & \textbf{92\%} & 62\% & 45\% & 67\% & 68\% & 67\% & 72\% \\
\textit{Pick Diverse Bottles} & 5\% & 3\% & 58\% & 36\% & 90\% & 91\% & 82\% & 70\% & \textbf{91\%} & \textbf{92\%} & 80\% & 85\% & 68\% & 68\% & 90\% & 91\% \\
\textit{Pick Dual Bottles} & 10\% & 6\% & 47\% & 36\% & 96\% & 90\% & 86\% & 86\% & 99\% & \textbf{100\%} & \textbf{100\%} & 96\% & 75\% & 91\% & \textbf{100\%} & 99\% \\
\textit{Place A2b Left} & 62\% & 60\% & 48\% & 49\% & 88\% & 79\% & 94\% & 88\% & 90\% & 87\% & \textbf{95\%} & \textbf{93\%} & 89\% & 82\% & 94\% & \textbf{93\%} \\
\textit{Place A2b Right} & 62\% & 57\% & 36\% & 36\% & 91\% & 87\% & 90\% & 92\% & 92\% & 89\% & 93\% & \textbf{99\%} & 82\% & 86\% & \textbf{99\%} & 96\% \\
\textit{Place Bread Basket} & 48\% & 56\% & 81\% & 71\% & 91\% & 94\% & 82\% & 82\% & 90\% & 91\% & 91\% & 93\% & 89\% & 77\% & \textbf{96\%} & \textbf{97\%} \\
\textit{Place Bread Skillet} & 38\% & 46\% & 77\% & 67\% & 86\% & 83\% & \textbf{94\%} & 90\% & 90\% & \textbf{96\%} & 90\% & 93\% & 92\% & 83\% & 93\% & 89\% \\
\textit{Place Burger Fries} & 66\% & 70\% & 94\% & 94\% & 98\% & 98\% & 98\% & 96\% & 97\% & \textbf{99\%} & 96\% & \textbf{99\%} & \textbf{100\%} & 91\% & 99\% & \textbf{99\%} \\
\textit{Place Can Basket} & 19\% & 25\% & 49\% & 52\% & 81\% & 76\% & 78\% & 74\% & \textbf{84\%} & \textbf{82\%} & 71\% & 69\% & 70\% & 74\% & 63\% & 56\% \\
\textit{Place Cans Plasticbox} & 40\% & 47\% & 97\% & 98\% & 98\% & 94\% & \textbf{100\%} & \textbf{100\%} & 99\% & 98\% & 99\% & 96\% & 96\% & 92\% & 99\% & 96\% \\
\textit{Place Container Plate} & 71\% & 78\% & 97\% & 95\% & \textbf{98\%} & 99\% & \textbf{98\%} & 96\% & \textbf{98\%} & \textbf{100\%} & 96\% & \textbf{100\%} & \textbf{98\%} & 97\% & \textbf{98\%} & 98\% \\
\textit{Place Dual Shoes} & 12\% & 7\% & 79\% & \textbf{88\%} & 93\% & 87\% & \textbf{96\%} & 84\% & 83\% & 81\% & 94\% & \textbf{88\%} & 75\% & 73\% & 82\% & 77\% \\
\textit{Place Empty Cup} & 75\% & 86\% & \textbf{100\%} & 98\% & 99\% & 98\% & 90\% & 90\% & 98\% & 99\% & \textbf{100\%} & \textbf{100\%} & 92\% & 93\% & \textbf{100\%} & \textbf{100\%} \\
\textit{Place Fan} & 25\% & 36\% & 80\% & 75\% & 91\% & 87\% & 92\% & 94\% & 84\% & 92\% & \textbf{96\%} & \textbf{96\%} & 83\% & 71\% & 94\% & 94\% \\
\textit{Place Mouse Pad} & 21\% & 26\% & 70\% & 70\% & 66\% & 68\% & 88\% & 90\% & 84\% & 86\% & 83\% & 89\% & 68\% & 68\% & \textbf{96\%} & \textbf{94\%} \\
\textit{Place Object Basket} & 43\% & 36\% & 44\% & 39\% & 81\% & 87\% & \textbf{90\%} & \textbf{92\%} & 85\% & 87\% & 89\% & 88\% & 89\% & 90\% & 82\% & 83\% \\
\textit{Place Object Scale} & 40\% & 49\% & 52\% & 74\% & 88\% & 85\% & 88\% & 80\% & 93\% & 89\% & 90\% & \textbf{97\%} & 86\% & 90\% & \textbf{95\%} & 93\% \\
\textit{Place Object Stand} & 74\% & 65\% & 86\% & 88\% & 98\% & 97\% & \textbf{100\%} & \textbf{98\%} & 97\% & 96\% & 90\% & 94\% & 88\% & 93\% & 90\% & 94\% \\
\textit{Place Phone Stand} & 49\% & 53\% & 88\% & 87\% & 87\% & 86\% & 82\% & 72\% & 75\% & 80\% & \textbf{97\%} & \textbf{99\%} & 56\% & 63\% & 96\% & 96\% \\
\textit{Place Shoe} & 57\% & 61\% & 96\% & 95\% & \textbf{99\%} & 97\% & 98\% & 96\% & 97\% & 99\% & 96\% & 99\% & 93\% & 89\% & \textbf{99\%} & \textbf{100\%} \\
\textit{Press Stapler} & 80\% & 70\% & 92\% & \textbf{98\%} & 93\% & \textbf{98\%} & 96\% & 96\% & 94\% & 90\% & 90\% & 97\% & \textbf{100\%} & \textbf{98\%} & 88\% & 90\% \\
\textit{Put Bottles Dustbin} & 12\% & 9\% & 74\% & 77\% & 81\% & 79\% & 72\% & 70\% & 85\% & \textbf{95\%} & 95\% & 90\% & 63\% & 59\% & \textbf{96\%} & 94\% \\
\textit{Put Object Cabinet} & 24\% & 15\% & 46\% & 48\% & 88\% & 71\% & 74\% & 74\% & 66\% & 76\% & \textbf{94\%} & \textbf{89\%} & 84\% & 85\% & 92\% & 86\% \\
\textit{Rotate Qrcode} & 47\% & 56\% & 34\% & 33\% & 89\% & 73\% & 90\% & 84\% & 84\% & 83\% & \textbf{93\%} & 89\% & 76\% & 73\% & 88\% & \textbf{91\%} \\
\textit{Scan Object} & 42\% & 38\% & 14\% & 36\% & 67\% & 66\% & 60\% & 64\% & 86\% & 79\% & 89\% & 92\% & 60\% & 50\% & \textbf{92\%} & \textbf{94\%} \\
\textit{Shake Bottle Horizontally} & 96\% & \textbf{100\%} & \textbf{100\%} & \textbf{100\%} & \textbf{100\%} & 98\% & \textbf{100\%} & 98\% & \textbf{100\%} & 99\% & \textbf{100\%} & \textbf{100\%} & \textbf{100\%} & 98\% & 98\% & 97\% \\
\textit{Shake Bottle} & 91\% & \textbf{100\%} & 99\% & \textbf{100\%} & \textbf{100\%} & 97\% & \textbf{100\%} & \textbf{100\%} & 99\% & 99\% & \textbf{100\%} & \textbf{100\%} & \textbf{100\%} & \textbf{100\%} & 99\% & 97\% \\
\textit{Stack Blocks Three} & 15\% & 16\% & 6\% & 10\% & 91\% & 95\% & 70\% & 78\% & 97\% & 95\% & 95\% & \textbf{97\%} & 85\% & 63\% & \textbf{99\%} & 95\% \\
\textit{Stack Blocks Two} & 48\% & 56\% & 92\% & 87\% & \textbf{100\%} & 98\% & \textbf{100\%} & 94\% & \textbf{100\%} & \textbf{100\%} & \textbf{100\%} & \textbf{100\%} & 95\% & 93\% & \textbf{100\%} & \textbf{100\%} \\
\textit{Stack Bowls Three} & 33\% & 35\% & 76\% & 86\% & 79\% & \textbf{87\%} & 70\% & 72\% & \textbf{88\%} & 82\% & 80\% & 81\% & 75\% & 79\% & 81\% & 83\% \\
\textit{Stack Bowls Two} & 78\% & 66\% & 96\% & 93\% & 98\% & \textbf{98\%} & 96\% & 92\% & 98\% & \textbf{98\%} & 92\% & \textbf{98\%} & \textbf{100\%} & 91\% & 95\% & 95\% \\
\textit{Stamp Seal} & 36\% & 23\% & 76\% & 82\% & 93\% & 92\% & \textbf{96\%} & \textbf{98\%} & 93\% & 95\% & 90\% & 94\% & 74\% & 81\% & 94\% & 96\% \\
\textit{Turn Switch} & 5\% & 6\% & 40\% & 61\% & \textbf{84\%} & 78\% & 82\% & \textbf{84\%} & 61\% & 72\% & 61\% & 59\% & 73\% & 66\% & 83\% & 75\% \\
\midrule
\textbf{\textit{Average (\%)}} & 42.98 & 43.84 & 72.88 & 72.84 & 88.66 & 87.02 & 86.36 & 85.04 & 89.76 & 90.68 & 91.88 & 91.78 & 82.40 & 80.80 & \textbf{92.94} & \textbf{92.14} \\
    \bottomrule
  \end{tabular}
\vspace{-0.3cm}
\end{adjustbox}
\end{table*}

\section{WorldArena Evaluation Details}
\label{sec:appx_experimental_details}

This section provides a detailed description of the WorldArena evaluation protocol, including metric definitions and the grouping strategy used to assess visual fidelity, physical consistency, and controllability.

\subsection{Metric Grouping and Aggregate Scoring}
\label{sec:appx_worldarena}

The WorldArena benchmark comprises 16 metrics, normalized to a $[0,100]$ scale where higher values indicate superior performance. The \textbf{EWMScore} is defined as the arithmetic mean of these 16 metrics. Detailed results for all metrics are presented in Table~\ref{tab:worldarena_full}.

The metrics are organized into six functional groups:
\begin{itemize}[leftmargin=2em, labelsep=0.45em, itemsep=1pt, topsep=2pt, parsep=0pt, partopsep=0pt]
    \item \textbf{Visual Quality}: frame-level realism and feature similarity (IQ, AQ, JEPA).
    \item \textbf{Motion Quality}: the presence and temporal smoothness of motion (DD, FS, MS).
    \item \textbf{Content Consistency}: stability of subjects, backgrounds, and photometric properties (SC, BC, PC).
    \item \textbf{Physics Adherence}: interaction plausibility and trajectory accuracy (IntQ, TA).
    \item \textbf{3D Accuracy}: depth consistency and perspectivity (DA, Per).
    \item \textbf{Controllability}: uniformity with instructions and action-conditioned variations (InsF, SA, AcF).
\end{itemize}

\begin{table}[t]
\centering
\caption{\textbf{Per-metric WorldArena results.}
All scores are reported on a $[0,100]$ scale.}
\label{tab:worldarena_full}
\renewcommand{\arraystretch}{1.05}
\resizebox{\linewidth}{!}{%
\begin{tabular}{l ccc ccc ccc cc cc ccc c}
\toprule
 & \multicolumn{3}{c}{\textbf{Visual Quality}} & \multicolumn{3}{c}{\textbf{Motion Quality}} & \multicolumn{3}{c}{\textbf{Content Consistency}} & \multicolumn{2}{c}{\textbf{Physics}} & \multicolumn{2}{c}{\textbf{3D Accuracy}} & \multicolumn{3}{c}{\textbf{Controllability}} & \\
\cmidrule(lr){2-4}\cmidrule(lr){5-7}\cmidrule(lr){8-10}\cmidrule(lr){11-12}\cmidrule(lr){13-14}\cmidrule(lr){15-17}
\textbf{Method} & IQ & AQ & JEPA & DD & FS & MS & SC & BC & PC & IntQ & TA & DA & Per & InsF & SA & AcF & \textbf{EWMScore} \\
\midrule
CogVideoX                       & 36.23             & 36.30             & 94.84             & 31.47             & 21.77             & 73.30             & 80.97             & 88.38             & 33.79             & 66.96             & 34.79             & 91.09             & 85.46             & 75.58             & 89.70             & 0.00              & 58.79 \\
Veo 3.1                         & \underline{65.57} & \textbf{48.79}    & 57.97             & 14.66             & 8.26              & 67.85             & 75.82             & \textbf{91.67}    & 37.52             & \underline{81.50} & 11.36             & 74.27             & \textbf{99.64}    & \textbf{97.14}    & 83.79             & 8.52              & 57.77 \\
Wan 2.6                         & \textbf{67.36}    & \underline{44.40} & 72.57             & 33.63             & 22.01             & \textbf{82.12}    & 73.15             & 84.29             & 34.57             & 73.16             & 12.18             & 75.43             & 93.94             & 89.96             & 88.09             & \underline{9.92}  & 59.80 \\
Cosmos-Predict 2.5 (text)       & 65.49             & 44.33             & 46.69             & 21.62             & 17.55             & 70.93             & 66.19             & 73.98             & 35.39             & 65.14             & 11.60             & 70.71             & 98.22             & 64.70             & 86.34             & \textbf{10.01}    & 53.06 \\
Cosmos-Predict 2.5 (action)     & 49.48             & 34.14             & 90.65             & 18.96             & 9.95              & 65.66             & 67.24             & 72.72             & \underline{45.21} & 60.46             & 27.49             & 90.35             & 84.68             & 60.54             & \underline{89.86} & 1.31              & 54.29 \\
IRASim                          & 41.85             & 32.86             & 93.72             & 21.60             & 11.66             & 64.31             & 72.82             & 78.69             & \textbf{45.70}    & 62.76             & 35.92             & 93.50             & 84.16             & 65.40             & 89.43             & 3.94              & 56.15 \\
Vidar                           & 41.75             & 39.87             & 61.10             & 21.05             & 17.03             & \underline{79.11} & 70.37             & 78.24             & 20.57             & 60.84             & 21.26             & 79.77             & 81.88             & 61.54             & 88.46             & 7.86              & 51.92 \\
Genie Envisioner                & 28.18             & 26.39             & 33.81             & 32.41             & 19.91             & 69.17             & 74.53             & 87.54             & 16.04             & 22.74             & 2.63              & 86.83             & 54.32             & 20.36             & 85.98             & 1.07              & 41.37 \\
Ctrl-World                      & 42.44             & 37.05             & 92.77             & \textbf{41.82}    & \textbf{33.57}    & 77.34             & \textbf{83.56}    & \underline{90.30} & 12.88             & 62.62             & 48.20             & 93.25             & 83.66             & 67.68             & 88.68             & 3.90              & 59.98 \\
ABot-PhysWorld (text)           & 61.03             & 41.83             & 90.36             & \underline{39.33} & \underline{29.22} & 76.48             & 80.56             & 89.02             & 20.52             & \textbf{81.96}    & 31.50             & 71.99             & \underline{98.94} & \underline{92.10} & 89.58             & 7.65              & \underline{62.63} \\
GigaWorld-1                     & 51.18             & 41.17             & \underline{96.77} & 30.52             & 18.64             & 68.33             & 80.97             & 86.43             & 28.11             & 75.10             & \underline{54.27} & \underline{98.44} & 95.60             & 82.14             & 89.42             & 0.28              & 62.34 \\
\midrule
\rowcolor{gray!15}
\textbf{FlowWAM (Ours)}         & 48.87             & 40.83             & \textbf{97.14}    & 30.41             & 20.47             & 71.65             & \underline{82.46} & 89.97             & 29.78             & 73.82             & \textbf{64.26}    & \textbf{98.97}    & 94.56             & 82.76             & \textbf{89.93}    & 3.50              & \textbf{63.71} \\
\bottomrule
\end{tabular}%
}
\end{table}

\subsection{Metric Definitions and Implementation}
\label{sec:appx_metric_defs}

We follow the official WorldArena metric definitions~\citep{worldarena}. Each raw metric is normalized to the common scoring range before computing EWMScore, so the descriptions below focus on what each metric measures and the main evaluator used by WorldArena.

\paragraph{Visual Quality.}
\begin{itemize}[leftmargin=2em, labelsep=0.45em, itemsep=1pt, topsep=2pt, parsep=0pt, partopsep=0pt]
    \item \textbf{IQ} (Image Quality) assesses the technical quality of individual frames using MUSIQ~\citep{musiq}. It targets no-reference distortions such as blur, overexposure, sensor noise, and compression artifacts, and averages the per-frame quality estimates over the video.
    \item \textbf{AQ} (Aesthetic Quality) evaluates whether frames have visually pleasing lighting, color composition, and perceptual appeal. WorldArena computes this score with the LAION aesthetic predictor, which maps each frame into an aesthetic feature space and averages the predicted frame scores.
    \item \textbf{JEPA} (JEPA Similarity) measures high-level spatiotemporal similarity between generated and reference videos. WorldArena extracts features with a pretrained V-JEPA encoder~\citep{vjepa} and computes a maximum mean discrepancy (MMD) score with a second-order polynomial kernel, so higher values indicate that the generated video distribution is closer to the ground-truth demonstration distribution.
\end{itemize}

\vspace{0.2cm}
\paragraph{Motion Quality.}
\begin{itemize}[leftmargin=2em, labelsep=0.45em, itemsep=1pt, topsep=2pt, parsep=0pt, partopsep=0pt]
    \item \textbf{DD} (Dynamic Degree) quantifies the strength of salient motion in the video. WorldArena estimates RAFT~\citep{raft} optical flow between adjacent frames, focuses on the top 5\% active pixels, and maps their average magnitude through a resolution-adaptive scoring function.
    \item \textbf{FS} (Flow Score) measures the overall motion intensity across the full video. It averages dense RAFT flow magnitudes over pixels and time, then normalizes the raw value so that videos with meaningful dynamic interaction receive higher scores.
    \item \textbf{MS} (Motion Smoothness) evaluates temporal coherence under motion. WorldArena uses a video frame interpolation model to reconstruct intermediate frames from neighboring frames, compares the interpolation with the real middle frame using SSIM, and weights the score by motion magnitude to avoid over-rewarding static videos.
\end{itemize}

\paragraph{Content Consistency.}
\begin{itemize}[leftmargin=2em, labelsep=0.45em, itemsep=1pt, topsep=2pt, parsep=0pt, partopsep=0pt]
    \item \textbf{SC} (Subject Consistency) measures whether foreground objects and robot parts remain semantically stable across frames. It compares DINO feature similarity between the current frame, the first frame, and the previous frame, with a dynamic-degree penalty so that nearly static videos cannot obtain artificially high consistency scores.
    \item \textbf{BC} (Background Consistency) assesses whether the scene background remains stable over time. It uses CLIP image features to compare each frame with the first and previous frames, and applies the same dynamic-degree adjustment to avoid rewarding static or frozen rollouts.
    \item \textbf{PC} (Photometric Consistency) evaluates pixel-level texture stability during motion. WorldArena warps frames forward and backward with optical flow, computes the average endpoint reconstruction error, and converts the reciprocal error into a normalized score with a dynamic-degree penalty.
\end{itemize}

\paragraph{Physics and 3D Accuracy.}
\begin{itemize}[leftmargin=2em, labelsep=0.45em, itemsep=1pt, topsep=2pt, parsep=0pt, partopsep=0pt]
    \item \textbf{IntQ} (Interaction Quality) measures whether robot-object interactions obey plausible physical behavior. WorldArena uses Qwen3-VL as a VLM judge to score visible contact, force transfer, friction, penetration, and object response on a 1--5 Likert scale before normalization.
    \item \textbf{TA} (Trajectory Accuracy) quantifies whether the generated robot-arm trajectory follows the ground-truth execution. It extracts robotic-arm bounding boxes with SAM3~\citep{sam3}, interpolates missing detections when necessary, and computes a normalized dynamic time warping (NDTW) alignment score between generated and reference trajectories.
    \item \textbf{DA} (Depth Accuracy) evaluates whether the generated rollout preserves the scene's spatial geometry. WorldArena estimates monocular depth for generated and reference frames, applies median-based scale alignment to handle depth-scale ambiguity, computes absolute relative depth error, and inverts the normalized error so that higher values indicate better geometry.
    \item \textbf{Per} (Perspectivity) assesses 3D plausibility beyond depth-map matching. A Qwen3-VL-based judge evaluates perspective cues such as scale changes with depth, occlusion relationships, lighting consistency, and camera-geometry stability on a normalized Likert scale.
\end{itemize}

\paragraph{Controllability.}
\begin{itemize}[leftmargin=2em, labelsep=0.45em, itemsep=1pt, topsep=2pt, parsep=0pt, partopsep=0pt]
    \item \textbf{InsF} (Instruction Following) evaluates whether the generated video follows the requested action type, target object, and final task state. WorldArena uses Qwen3-VL as a VLM judge with a normalized 1--5 scale, and the rubric penalizes failures such as wrong objects, wrong manipulation type, or hallucinated human hands in place of robot arms.
    \item \textbf{SA} (Semantic Alignment) measures semantic agreement between generated and reference executions. WorldArena first uses Qwen2.5-VL to produce structured descriptions of both videos, then computes CLIP text-feature similarity between those descriptions.
    \item \textbf{AcF} (Action Following) evaluates whether a model responds differently to different action instructions instead of collapsing to repetitive rollouts. For a shared initial frame, WorldArena generates multiple distinct action instructions, extracts global CLIP features from the resulting videos, and scores the average pairwise feature dissimilarity.
\end{itemize}

\subsection{Performance Analysis on WorldArena}
\label{sec:appx_wa_analysis}

The comprehensive metric profile in Table~\ref{tab:worldarena_full} reveals that FlowWAM's performance gains are primarily driven by metrics requiring action-grounded prediction. Specifically, FlowWAM achieves state-of-the-art results in \textbf{Trajectory Accuracy}, \textbf{Depth Accuracy}, and \textbf{Semantic Alignment}. These results suggest that flow conditioning provides a spatially aligned signal that guides the video generator in accurately modeling object and robot dynamics, rather than merely improving generic image aesthetics. While FlowWAM maintains competitive performance in subject and background consistency, its core strength lies in its ability to generate physically plausible and reliable motion rollouts.

\section{Real-World Experimental Setup}
\label{sec:appx_realworld_setup}

This section provides additional details regarding the real-world experimental setup, including the robot platforms, camera configurations, and the task suite used for evaluation.

\begin{figure}[ht]
    \centering
    \includegraphics[width=\linewidth]{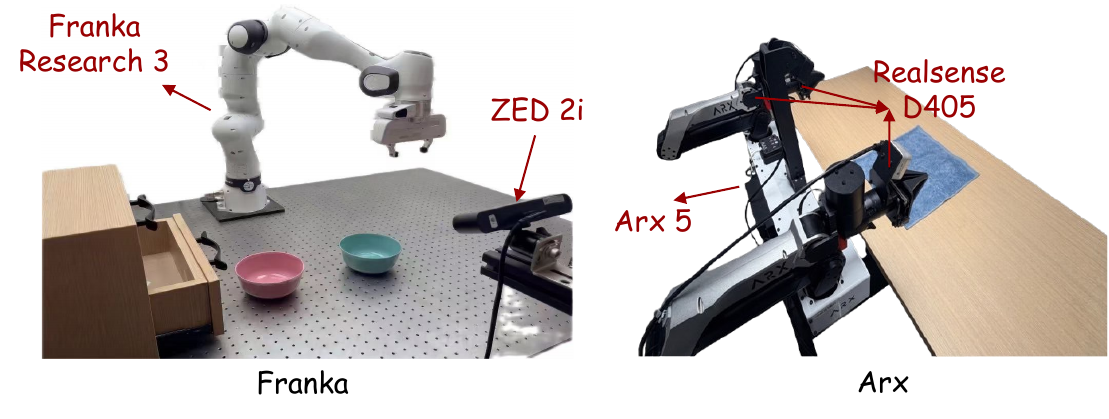}
    \caption{
        \textbf{Real-world experimental setup.}
        We evaluate FlowWAM on two physical platforms: a single-arm Franka Research~3 robot and a bimanual ARX~5 robot.
    }
    \label{fig:appx_realworld_setup}
\end{figure}

\subsection{Robot Platforms and Sensors}
We use two robot platforms to evaluate the cross-embodiment transferability of FlowWAM:
\begin{itemize}[leftmargin=2em, labelsep=0.45em, itemsep=1pt, topsep=2pt, parsep=0pt, partopsep=0pt]
    \item \textbf{Franka Research~3}: a single-arm manipulation platform controlled through FrankaPy~\citep{frankapy}, equipped with a parallel-jaw gripper and one ZED camera for workspace observation, and used for precision pick-and-place and contact-rich object manipulation.
    \item \textbf{ARX~5}: a bimanual robot platform using an OCS2-based optimal-control stack, equipped with three Intel RealSense D405 cameras, including one front camera for the global workspace view and two wrist cameras for close-range manipulation, and used for tasks that require coordinated motion between two arms.
\end{itemize}

\subsection{Task Suite and Data Collection}
The real-world evaluation suite consists of seven tasks designed to test spatial precision, object interaction, and bimanual coordination:
\begin{itemize}[leftmargin=2em, labelsep=0.45em, itemsep=1pt, topsep=2pt, parsep=0pt, partopsep=0pt]
    \item \textbf{Stack Bowls} (Franka): pick up one bowl and place it stably on top of another bowl.
    \item \textbf{Place in Drawer} (Franka): grasp an object and insert it into an open drawer while avoiding collisions with the drawer boundary.
    \item \textbf{Put in Plate} (Franka): move a target object into a plate, requiring accurate placement relative to a shallow container.
    \item \textbf{Place Two Cups} (Franka): sequentially manipulate two cups and place them at the desired target locations.
    \item \textbf{Fold Towel} (ARX): use both arms to grasp and fold a deformable towel.
    \item \textbf{Stack Bowls} (ARX): coordinate two arms to grasp, transfer, and stack bowls in a stable configuration.
    \item \textbf{Clean Plate} (ARX): execute a bimanual cleaning motion over a plate, requiring coordinated arm motion over an extended horizon.
\end{itemize}
For each task, all methods are trained on the same 100 teleoperated demonstrations and evaluated over 10 trials with randomized object poses.

\section{Visualization Results}
\label{sec:appx_visualization}

This appendix provides qualitative visualizations that complement the quantitative evaluations in Section~\ref{sec:experiments}.
The goal is to make the behavior of FlowWAM directly inspectable. Beyond aggregate metrics, the generated rollouts should preserve the intended robot motion, object interaction, and scene geometry over time.

\subsection{RoboTwin Policy Rollouts}
\label{sec:appx_robotwin_vis}

\begin{figure}[t]
    \centering
    \includegraphics[width=\linewidth]{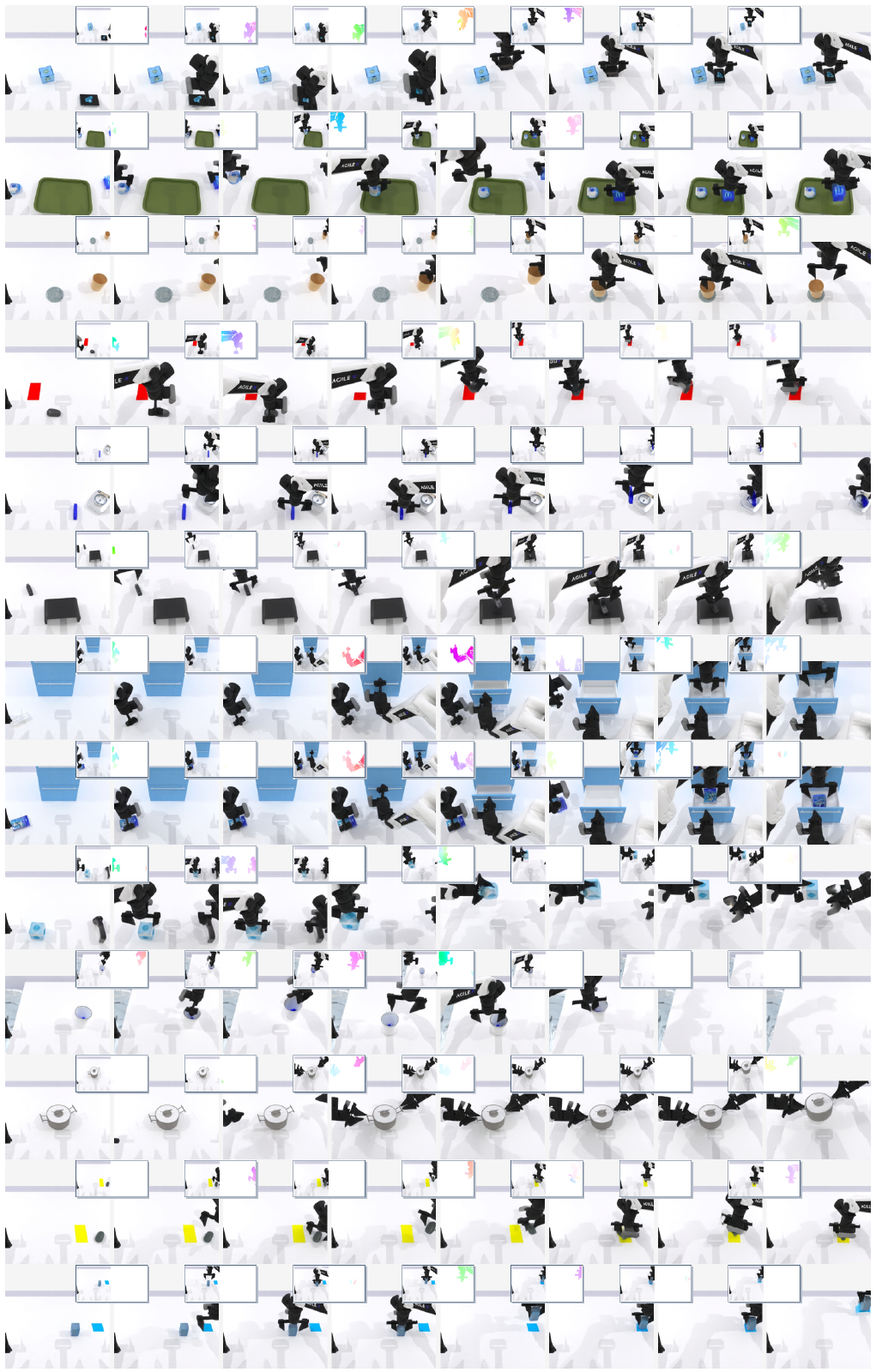}
    \caption{
        Qualitative policy rollouts on RoboTwin~2.0 \textit{Clean} tasks.
        For each task, the small panels show FlowWAM's predicted RGB frames and the corresponding optical-flow plan, and the large panels show the simulation rollout after the action expert decodes the predicted flow into actions.
    }
    \label{fig:appx_robotwin_clean}
\end{figure}

\begin{figure}[t]
    \centering
    \includegraphics[width=\linewidth]{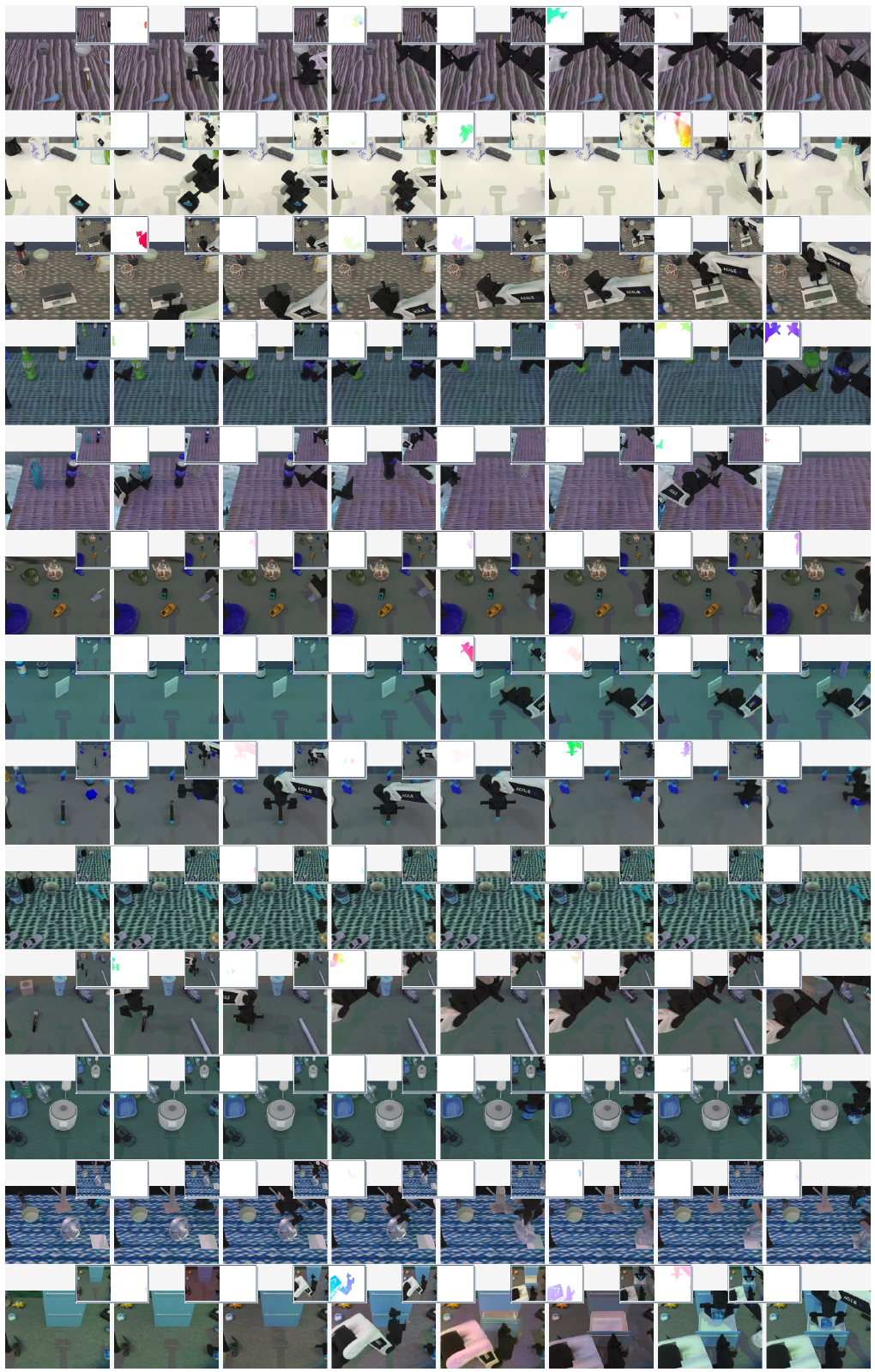}
    \caption{
        Qualitative policy rollouts on RoboTwin~2.0 \textit{Random} tasks.
        Compared with the \textit{Clean} setting in Figure~\ref{fig:appx_robotwin_clean}, the scenes here include randomized object pose, distractors, lighting, and background.
    }
    \label{fig:appx_robotwin_random}
\end{figure}

Figures~\ref{fig:appx_robotwin_clean} and~\ref{fig:appx_robotwin_random} extend the policy mode visualizations in Figure~\ref{fig:qualitative_main} to a broader set of RoboTwin~2.0~\citep{robotwin2} tasks under both the \textit{Clean} and \textit{Random} settings.
For each task, we show the predicted RGB rollout together with its corresponding optical-flow plan, and the executed simulation rollout decoded by the action expert.
The \textit{Clean} examples (Figure~\ref{fig:appx_robotwin_clean}) cover diverse bimanual manipulation skills under the canonical layout, while the \textit{Random} examples (Figure~\ref{fig:appx_robotwin_random}) probe the same skills under randomized object pose, distractor placement, lighting, and background.
Across both settings, the predicted flow concentrates on the moving embodiment and traces a coherent goal-directed trajectory, while the RGB rollout preserves the surrounding object context. The decoded actions follow this robot-motion plan to complete the task, and the qualitative behavior remains stable as the scene is randomized, mirroring the quantitative robustness reported in Table~\ref{tab:robotwin-full}.

\subsection{Real-World Policy Rollouts}
\label{sec:appx_realworld_policy}

\begin{figure}[t]
    \centering
    \includegraphics[width=\linewidth]{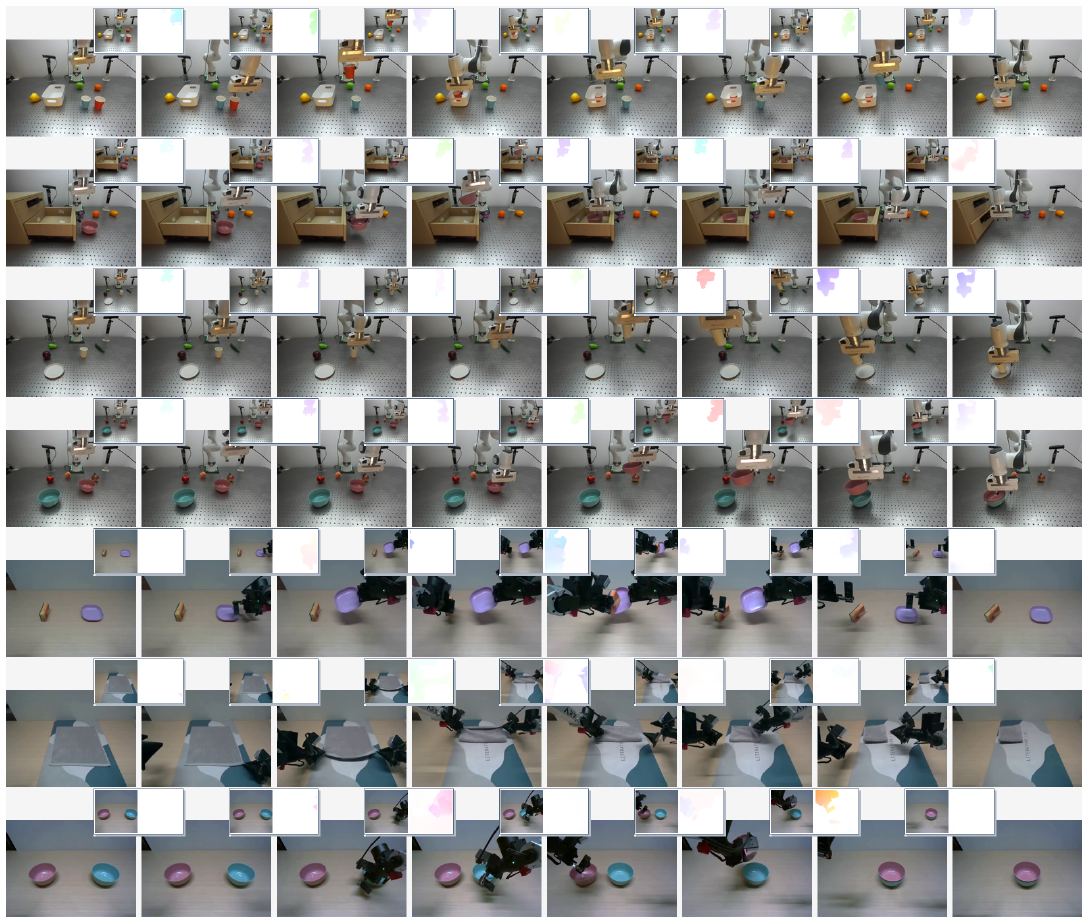}
    \caption{
        Real-world execution examples in policy mode across single-arm and dual-arm platforms.
        The first four rows correspond to single-arm Franka tasks, and the last three rows correspond to dual-arm ARX tasks.
        Within each task, FlowWAM's predicted RGB frames and optical-flow plans are paired with the corresponding physical executions.
        The qualitative alignment between the predicted motion fields and the realized robot trajectories shows how the video-native flow representation connects model prediction with executable control.
    }
    \label{fig:appx_realworld_policy}
\end{figure}

Figure~\ref{fig:appx_realworld_policy} visualizes FlowWAM's policy mode behavior on the real-world manipulation suite introduced in Section~\ref{sec:exp_realworld}, covering both the single-arm Franka platform and the bimanual ARX platform.
The first four rows show single-arm Franka tasks (\textit{Stack Bowls}, \textit{Place in Drawer}, \textit{Put in Plate}, \textit{Place Two Cups}), which stress spatial precision and object contact.
The last three rows show dual-arm ARX tasks (\textit{Fold Towel}, \textit{Stack Bowls}, \textit{Clean Plate}), which additionally require bimanual coordination and longer-horizon motion.
For each task, we juxtapose the model-side prediction (predicted RGB rollout and the corresponding optical-flow plan) with the realized robot execution after the predicted flow is decoded into continuous actions by the action expert.
This side-by-side layout makes the role of flow directly inspectable across both single-arm and bimanual settings: the predicted flow provides a dense motion plan in image space, while the executed rollout shows whether that plan translates into the intended end-effector motion and object interaction on the physical robot.

\subsection{World Model Visualization}
\label{sec:appx_world_model_visualization}

\begin{figure}[t]
    \centering
    \includegraphics[width=0.95\linewidth]{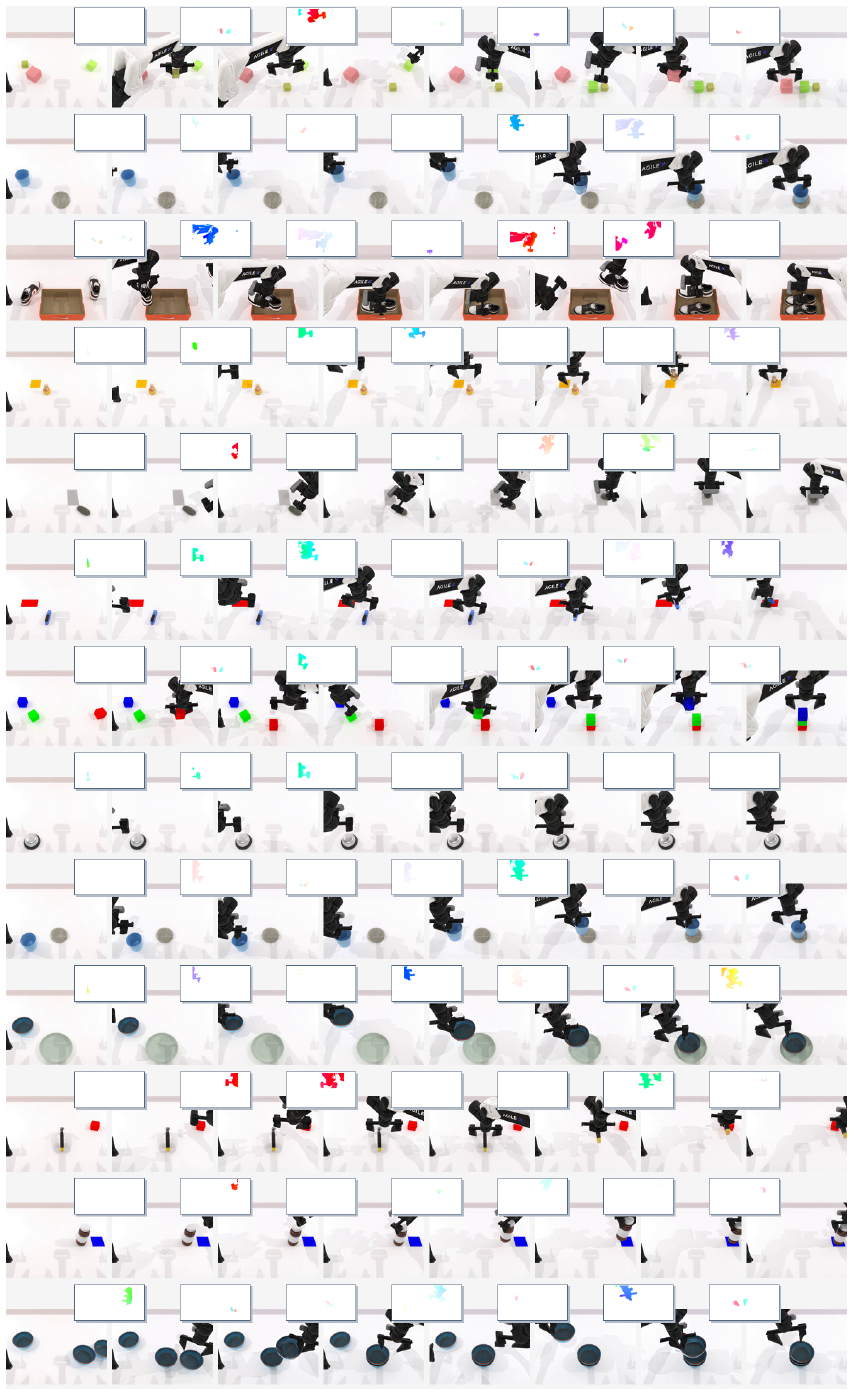}
    \caption{
        Qualitative visualization of FlowWAM's action-conditioned world model.
        The examples show how the predicted future frames follow the requested manipulation dynamics while preserving object layout and scene consistency.
    }
    \label{fig:appx_world_model_vis}
\end{figure}

Figure~\ref{fig:appx_world_model_vis} visualizes representative action-conditioned world-model predictions.
Given the initial observation and the target action condition, FlowWAM generates future RGB frames together with the corresponding motion structure.
The visualization illustrates how the dense optical-flow representation provides spatially grounded guidance for video prediction, helping the model maintain coherent robot-object motion and reduce action drift across the rollout.

\section{Broader Impacts}
\label{sec:appx_broader_impacts}

FlowWAM studies flow-based action representations for embodied world models in simulated and lab-scale manipulation, with potential benefits for more reusable and data-efficient robot learning.
We do not anticipate direct societal impacts beyond those generally associated with foundational robot-learning research.



\end{document}